\documentclass{article}

\usepackage{hyperref}

\usepackage[arxiv]{icml2020}

\usepackage{url}
\usepackage{multicol}
\usepackage{amsmath}
\usepackage{amsthm}
\usepackage{booktabs}
\usepackage{tabularx}
\usepackage{xcolor}
\usepackage{natbib}
\usepackage{multirow}
\usepackage{verbatim}
\usepackage{siunitx}
\usepackage{algorithm}
\usepackage{wrapfig}
\usepackage{graphicx}
\usepackage{subcaption}
\graphicspath{ {figures/} }

\usepackage{xr}
\usepackage{amsmath,amsfonts,bm}

\def\eqref#1{equation~\ref{#1}}

\def\1{\bm{1}}

\DeclareMathAlphabet{\mathsfit}{\encodingdefault}{\sfdefault}{m}{sl}
\SetMathAlphabet{\mathsfit}{bold}{\encodingdefault}{\sfdefault}{bx}{n}

\def\gM{{\mathcal{M}}}
\def\gN{{\mathcal{N}}}

\newcommand{\R}{\mathbb{R}}

\newcommand{\grad}{\nabla}
\newcommand{\loss}{\ell}

\theoremstyle{definition}
\newtheorem{defn}{Definition}

\newcommand{\deepobs}{\text{\textsc{DeepOBS}}}
\newcommand{\sgd}{\text{\textsc{SGD}}}
\newcommand{\momentum}{\text{\textsc{Momentum}}}
\newcommand{\nesterov}{\text{\textsc{Nesterov}}}
\newcommand{\adam}{\text{\textsc{Adam}}}
\newcommand{\nadam}{\text{\textsc{NAdam}}}
\newcommand{\rmsprop}{\text{\textsc{RMSProp}}}
\newcommand{\rmsterov}{\text{\textsc{RMSterov}}}

\newcommand{\N}{\mathbb{N}}
\newcommand{\xmin}{\theta_{\star}}

\newcommand{\optarraystretch}{1}
\newcommand{\sgdheader}{$\sgd(H_t, \eta_t)$}
\newcommand{\sgdbody}{
$
\begin{aligned}
    \theta_{t+1} = \theta_{t} - \eta_t \grad \loss(\theta_t)
\end{aligned}
$
}
\newcommand{\momheader}{$\momentum(H_t, \eta_t, \gamma)$}
\newcommand{\mombody}{
$
\begin{aligned}
    &v_0 = 0  \\
    &v_{t+1} = \gamma v_{t} + \grad \loss(\theta_t)\\
    &\theta_{t+1} = \theta_{t} - \eta_t v_{t+1}
\end{aligned}
$
}
\newcommand{\nesterovheader}{$\nesterov(H_t, \eta_t, \gamma)$}
\newcommand{\nesterovbody}{
$
\begin{aligned}
    &v_0 = 0  \\
    &v_{t+1} = \gamma v_{t} + \grad \loss(\theta_t) \\
    &\theta_{t+1} = \theta_{t} - \eta_t \left( \gamma v_{t+1} + \grad \loss(\theta_{t}) \right)
\end{aligned}
$
}
\newcommand{\rmspropheader}{$\rmsprop(H_t, \eta_t, \gamma, \rho, \epsilon)$}
\newcommand{\rmspropbody}{
$
\begin{aligned}
    &v_0 = 1 \text{,} \, m_0 = 0 \\
    &v_{t+1} = \rho v_{t} + (1 - \rho) \grad \loss(\theta_t)^2 \\
    &m_{t+1} = \gamma m_{t} + \frac{\eta_t}{\sqrt{v_{t+1} + \epsilon}}\grad \loss(\theta_t)\\
    &\theta_{t+1} = \theta_{t} - m_{t+1}
\end{aligned}
$
}

\newcommand{\adamheader}{$\adam(H_t, \alpha_t, \beta_1, \beta_2, \epsilon)$}
\newcommand{\adambody}{
$
\begin{aligned}
    &m_0 = 0 \text{,} \, v_0 = 0 \\
    &m_{t+1} = \beta_1 m_{t} + (1 - \beta_1) \grad \loss (\theta_t) &\\
    &v_{t+1} = \beta_2 v_{t} + (1 - \beta_2) \grad \loss(\theta_t)^2\\
    &b_{t+1} = \frac{\sqrt{1 - \beta_2^{t+1}}}{1 - \beta_1^{t+1}}\\
    &\theta_{t+1} = \theta_{t} - \alpha_t \frac{m_{t+1}}{\sqrt{v_{t+1}} + \epsilon} b_{t+1}
\end{aligned}
$
}
\newcommand{\nadamheader}{$\nadam(H_t, \alpha_t, \beta_1, \beta_2, \epsilon)$}
\newcommand{\nadambody}{
$
\begin{aligned}
    &m_0 = 0 \text{,} \, v_0 = 0 \\
    &m_{t+1} = \beta_1 m_{t} + (1 - \beta_1) \grad \loss (\theta_t) &\\
    &v_{t+1} = \beta_2 v_{t} + (1 - \beta_2) \grad \loss (\theta_t)^2\\
    &b_{t+1} = \frac{\sqrt{1 - \beta_2^{t+1}}}{1 - \beta_1^{t+1}}\\
    &\theta_{t+1} = \theta_{t} - \alpha_t \frac{\beta_1 m_{t+1} + (1 - \beta_1) \grad \loss (\theta_t)}{\sqrt{v_{t+1}} + \epsilon} b_{t+1}
\end{aligned}
$
}

\icmltitlerunning{On Empirical Comparisons of Optimizers for Deep Learning}

\begin{document}

\twocolumn[
\icmltitle{On Empirical Comparisons of Optimizers for Deep Learning}



\icmlsetsymbol{equal}{*}

\begin{icmlauthorlist}
\icmlauthor{Dami Choi}{gb,to}
\icmlauthor{Christopher J. Shallue}{gb}
\icmlauthor{Zachary Nado}{gb}
\icmlauthor{Jaehoon Lee}{gb}
\icmlauthor{Chris J. Maddison}{dm,ifas}
\icmlauthor{George E. Dahl}{gb}
\end{icmlauthorlist}

\icmlaffiliation{gb}{Google Research, Brain Team}
\icmlaffiliation{dm}{DeepMind}
\icmlaffiliation{ifas}{Institute for Advanced Study, Princeton NJ}
\icmlaffiliation{to}{Vector Institute and University of Toronto}

\icmlcorrespondingauthor{Dami Choi}{choidami@cs.toronto.edu}

\icmlkeywords{Machine Learning, ICML}

\vskip 0.3in
]



\printAffiliationsAndNotice{}  

\begin{abstract}
Selecting an optimizer is a central step in the contemporary deep learning pipeline. In this paper, we demonstrate the sensitivity of optimizer comparisons to the hyperparameter tuning protocol.
Our findings suggest that the hyperparameter search space may be the single most important factor explaining the rankings obtained by recent empirical comparisons in the literature.
In fact, we show that these results can be contradicted when hyperparameter search spaces are changed. As tuning effort grows without bound, more general optimizers should never underperform the ones they can approximate (i.e., Adam should never perform worse than momentum), but recent attempts to compare optimizers either assume these inclusion relationships are not practically relevant or restrict the hyperparameters in ways that break the inclusions. In our experiments, we find that inclusion relationships between optimizers matter in practice and always predict optimizer comparisons. In particular, we find that the popular adaptive gradient methods never underperform momentum or gradient descent. We also report practical tips around tuning often ignored hyperparameters of adaptive gradient methods and raise concerns about fairly benchmarking optimizers for neural network training.
\end{abstract}

\section{Introduction}
\label{sec:introduction}

The optimization algorithm chosen by a deep learning practitioner determines the training speed and the final predictive performance of their model. To date, there is no theory that adequately explains how to make this choice. Instead, our community relies on empirical studies \citep{wilson2017marginal} and benchmarking \citep{schneider2019deepobs}. Indeed, it is the de facto standard that papers introducing new optimizers report extensive comparisons across a large number of workloads. Therefore, to maximize scientific progress, we must have confidence in our ability to make empirical comparisons between optimization algorithms.

Although there is no theory guiding us when comparing optimizers, the popular first-order optimizers form a natural inclusion hierarchy. For example, \adam{} \citep{kingma2014adam} and \rmsprop{} \citep{tieleman2012lecture} can approximately simulate \momentum{} \citep{polyak1964some} if the $\epsilon$ term in the denominator of their parameter updates is allowed to grow very large. However, these relationships may not matter in practice. For example, the settings of \adam{}'s hyperparameters that allow it to match the performance of \momentum{} may be too difficult to find (for instance, they may be infinite).

In this paper, we demonstrate two important and interrelated points about empirical comparisons of neural network optimizers. First, we show that inclusion relationships between optimizers actually matter in practice; in our experiments, more general optimizers \emph{never} underperform special cases. Despite conventional wisdom \citep{wilson2017marginal, balles2017dissecting}, we find that when carefully tuned, \adam{} and other adaptive gradient methods never underperform \momentum{} or \sgd{}. Second, we demonstrate the sensitivity of optimizer comparisons to the hyperparameter tuning protocol. By comparing to previous experimental evaluations, we show how easy it is to change optimizer rankings on a given workload (model and dataset pair) by changing the hyperparameter tuning protocol, with optimizer rankings stabilizing according to inclusion relationships as we spend more and more effort tuning. Our findings raise serious questions about the practical relevance of conclusions drawn from these sorts of empirical comparisons.

The remainder of this paper is structured as follows. In Section \ref{sec:relatedwork}, we review related work, focusing on papers that make explicit claims about optimizer comparisons in deep learning and application papers that provide evidence about the tuning protocols of practitioners. We develop our definition of first-order optimizers in Section \ref{sec:optimizerdefition} along with a notion of inclusion relationships between optimizers. We present our experimental results in Section~\ref{sec:experiments}. Despite thorny methodological issues over how to avoid biases in comparisons due to search spaces that favor one optimizer over another, we believe that our experimental methodology is an acceptable compromise and has substantial practical relevance.
Among other results, we show that the inclusion hierarchy of update rules is almost entirely predictive of optimizer comparisons. In particular, \nadam{} \citep{dozat2016incorporating} achieves the best top-1 validation accuracy on ResNet-50 on ImageNet in our experiments. The 77.1\% we obtain with \nadam, although not as good as the 77.6\% obtained using learned data augmentation by \citet{cubuk2018autoaugment}, is better than the best existing published results using any of the more standard pre-processing pipelines (76.5\%, due to \citet{goyal2017accurate} using \momentum{}).

\section{Background and Related Work}
\label{sec:relatedwork}

Our work was inspired by the recent studies of neural network optimizers by \citet{wilson2017marginal} and \citet{schneider2019deepobs}.
\citet{wilson2017marginal} constructed a simple classification problem in which adaptive gradient methods (e.g. \adam{}) converge to provably worse solutions than standard gradient methods. However, crucially, their analysis ignored the $\epsilon$ parameter in the denominator of some adaptive gradient methods. \citet{wilson2017marginal} also presented experiments in which \adam{} produced worse validation accuracy than \sgd{} across \textit{all} deep learning workloads considered. However, they only tuned over the learning rate and learning rate decay scheme in their experiments, leaving all other parameters of \adam{} at fixed default values.
Despite these findings, adaptive gradient methods continue to be popular since the work of \citet{wilson2017marginal}.
\citet{schneider2019deepobs} presented a benchmark suite (\deepobs) for deep learning optimizers and reported that there was no single best optimizer across the workloads they considered. Yet \citet{schneider2019deepobs} only tuned the learning rate of each optimizer and left all other hyperparameters at some fixed default values.

As we discuss in Section~\ref{sec:reconcilingdisagreements}, the choices of hyperparameter tuning protocols in  \citet{wilson2017marginal} and \citet{schneider2019deepobs} may be the most important factor preventing their results from being relevant to practical choices about which optimizer to use.
Hyperparameter tuning is a crucial step of the deep learning pipeline \citep{bergstra2012random, SnoekEtAl2012bayesopt,  sutskever2013importance, smith2018disciplinedapproach}, so it is critical for papers studying optimizers to match as closely as possible the tuning protocols of an ideal practitioner. Yet, tuning protocols often differ between works studying neural network optimizers and works concerned with training neural networks to solve specific problems. %

Recent papers that study or introduce optimization algorithms tend to compare to \adam{} and \rmsprop{} without tuning their respective $\epsilon$ hyperparameters (see Table~\ref{table:updaterules} for notations), presumably to simplify their experiments.
It is standard to leave $\epsilon$ at the common default value of $10^{-8}$ for \adam{} and $10^{-10}$ for \rmsprop{} \citep{tieleman2012lecture, kingma2014adam, dozat2016incorporating, balles2017dissecting, loshchilov2017fixing, zou2018convergence, ma2018quasi, bernstein2018signsgd, chen2019namsg, zou2019sufficient}. Others do not even report the value of $\epsilon$ used \citep{balles2017dissecting, zhang2017yellowfin, keskar2017improving, chen2018convergence, zhou2018adashift, aitchison2018unified, reddi2019convergence, luo2019adaptive}. There are exceptions. \citet{YOGI} and \citet{radam} considered $\epsilon$ values orders of magnitude larger than the standard default. However, the experiments in both papers gave only a limited consideration to $\epsilon$, testing at most two values while tuning \adam{}. \citet{de2018convergence} is the only work we found that considered a broad range of values for $\epsilon$. Both \citet{YOGI} and \citet{de2018convergence} found that non-default values of $\epsilon$ outperformed the default.

While it is also extremely common in applications to use a default value of $\epsilon$, some notable papers tuned $\epsilon$ and selected values up to eight orders of magnitude away from the common defaults. \citet{szegedy1512sergey} used $\epsilon=1$ for \rmsprop{}; \citet{liu2019roberta} reported that their results were sensitive to $\epsilon$ and set  $\epsilon=10^{-6}$ for \adam{}; \citet{tan2019mnasnet} and \citet{tan2019efficientnet} set $\epsilon = 10^{-3}$ for \rmsprop{}, the latter achieving state-of-the-art ImageNet top-1 accuracy. In reinforcement learning, \citet{rainbow} set $\epsilon = \num{1.5e-4}$.

Despite being introduced solely to prevent division by zero\footnote{TensorFlow currently refers to $\epsilon$ as ``a small constant for numerical stability''; \url{https://www.tensorflow.org/versions/r1.15/api_docs/python/tf/train/AdamOptimizer}.}
, \adam{}'s $\epsilon$ can be interpreted in ways that suggest the optimal choice is problem-dependent. If \adam{} is interpreted as an empirical, diagonal approximation to natural gradient descent \citep{kingma2014adam}, $\epsilon$ can be viewed as a multi-purpose damping term whose role is to improve the conditioning of the Fisher, in analogy to the approximate second-order method considered by \citet{becker1988improving}. We can also view $\epsilon$ as setting a trust region radius \citep{kfac, adolphs2019ellipsoidal} and controlling an interpolation between momentum and diagonal natural gradient descent, by either diminishing or increasing the effect of $v_t$ on the update direction. Under either interpretation, the best value for $\epsilon$ will be problem-dependent and likely benefit from tuning.

\section{What is an optimizer?}
\label{sec:optimizerdefition}

\begin{table*}[t!]
{
\caption{Update rules considered in this work. $\sgd$ is due to \cite{robbins1951stochastic}, $\momentum$ to \cite{polyak1964some}, $\nesterov$ to \cite{nesterov1983method}, $\rmsprop$ to \cite{tieleman2012lecture}, and $\nadam$ to \cite{dozat2016incorporating}. All operations are taken component-wise for vectors. In particular, for $x \in \R^d$, $x^2$ is a component-wise power function. }
\label{table:updaterules}
\def\arraystretch{\optarraystretch}

\rule{\textwidth}{1pt}

\vspace{0.5\baselineskip}

\begin{tabularx}{0.5\textwidth}{X}
    \sgdheader \\
    \cmidrule(lr){1-1} 
    \sgdbody \\
    \\
    \momheader  \\
    \cmidrule(lr){1-1}
    \mombody  \\
    \\
    \nesterovheader \\
    \cmidrule(lr){1-1}
    \nesterovbody \\
    \\
    \rmspropheader\\
    \cmidrule(lr){1-1}
    \rmspropbody\\
\end{tabularx}
\begin{tabularx}{0.5\textwidth}{X}
    \adamheader\\
    \cmidrule(lr){1-1}
    \adambody\\
    \\
    \nadamheader\\
    \cmidrule(lr){1-1}
    \nadambody\\
\end{tabularx}
}

\vspace{0.25\baselineskip}

\rule{\textwidth}{1pt}
\end{table*}

Optimization algorithms are typically defined by their update rule, which is controlled by hyperparameters that determine its behavior (e.g. the learning rate). 
Consider a differentiable loss function $\loss : \R^d \to \R$ whose vector of first partial derivatives is given by $\grad \loss(\theta)$ (more generally, $\grad \loss(\theta)$ might be a stochastic estimate of the true gradient). In our context, $\loss$ represents the loss function computed over an entire dataset by a neural network and $\theta \in \R^d$ represents the vector of model parameters. The optimization problem is to find a point that (at least locally) minimizes $\loss$. First-order iterative methods for this problem \citep{nesterov2018lectures} construct a sequence $\theta_t$ of iterates converging to a local minimum $\xmin$ using queries to $\loss$ and $\grad \loss$. The sequence $\theta_t$ is constructed by an update rule $\gM$, which determines the next iterate $\theta_{t+1}$ from the history $H_t$ of previous iterates along with their function and gradient values, $H_t = \{\theta_s, \grad \loss(\theta_s), \loss(\theta_s)\}_{s=0}^t$, and a setting of hyperparameters $\phi : \N \to \R^n$. Given an initial parameter value $\theta_0 \in \R^d$, the sequence of points visited by an optimizer with update rule $\gM$ is given by,
\begin{align*}
\theta_{t+1} = \gM(H_t, \phi_t).
\end{align*}
The stochastic gradient descent algorithm \citep[\sgd ;][]{robbins1951stochastic} is one of the simplest such methods used for training neural networks. $\sgd$ is initialized with $\theta_0 \in \R^d$, and its hyperparameter is a learning rate schedule $\eta: \N \to (0, \infty)$. The \sgd{} update rule is given by $\sgd(H_t, \eta_t) = \theta_t - \eta_t \grad \loss(\theta_t)$. The $\momentum$ method due to \citet{polyak1964some} generalizes the \sgd{} method by linearly combining the gradient direction with a constant multiple of the previous parameter update. Its hyperparameters are a learning rate schedule $\eta : \N \to (0, \infty)$ and a momentum parameter $\gamma \in [0, \infty)$,
\begin{align*}
\label{eq:momentum} \momentum(H_t, \eta_t, \gamma) = \theta_{t} - \eta_t \grad \loss(\theta_{t})  + \gamma (\theta_{t} - \theta_{t-1}).
\end{align*}
There has been an explosion of novel first-order methods in deep learning, all of which fall into this standard first-order scheme. In Table \ref{table:updaterules} we list the first-order update rules considered in this paper.

The difference between optimizers is entirely captured by the choice of update rule $\gM$ and hyperparameters $\phi$. Since the roles of optimizer hyperparameters on neural network loss functions are not well-understood, most practitioners tune a subset of the hyperparameters to maximize performance over a validation set, while leaving some hyperparameters at fixed default values. The choice of which hyperparameters to tune determines an \emph{effective} family of update rules, and this family is the critical object from a practitioners perspective. Thus, in analogy to (overloaded) function declarations in C++, we define an \emph{optimizer} by an update rule ``signature,'' the update rule name together with the free hyperparameter arguments. For example, in this definition $\momentum(\cdot, \eta_t, \gamma)$ is not the same optimizer as $\momentum(\cdot, \eta_t, 0.9)$, because the latter has two free hyperparameters while the former only has one. \adam{} with the default $\epsilon$ is ``different'' from \adam{} with tuned $\epsilon$.

\subsection{The taxonomy of first-order methods}\label{sec:taxonomy}

The basic observation of this section is that some optimizers can approximately simulate others (i.e., optimizer A might be able to approximately simulate the trajectory of optimizer B for any particular setting of B's hyperparameters). This is important knowledge because, as a hyperparameter tuning protocol approaches optimality, a more expressive optimizer can never underperform any of its specializations.
To capture the concept of one optimizer approximating another, we define the following inclusion relationship between optimizers.

\begin{defn}[Inclusion relationship] 
\label{def:inclusion}
Let $\gM, \gN$ be update rules for use in a first-order optimization method. $\gM$ is a subset or specialization of $\gN$, if for all $\phi : \N \to \R^n$, there exists a sequence $\psi^i : \N \to \R^m$, such that for all $t \in [0, \infty)$ and histories $H_t$,
\[\lim_{i \to \infty} \gN(H_t, \psi_t^{i}) = \gM(H_t, \phi_t)\]
This is denoted $\gM \subseteq \gN$, with equality $\gM = \gN$ iff $\gM \subseteq \gN$ and $\gN \subseteq \gM$.
\end{defn}
Evidently $\sgd \subseteq \momentum$, since $\sgd(H_t, \eta_t) =  \momentum(H_t, \eta_t, 0)$. Many well-known optimizers fall naturally into this taxonomy. In particular, we consider $\rmsprop$ with momentum \citep{tieleman2012lecture}, $\adam$ \citep{kingma2014adam} and $\nadam$ \citep{dozat2016incorporating} (see Table \ref{table:updaterules}) and show the following inclusions in the appendix.
\begin{equation*}
\begin{aligned}
    \sgd &\subseteq \momentum \subseteq \rmsprop\\
    \sgd &\subseteq \momentum \subseteq \adam\\
    \sgd &\subseteq \nesterov \subseteq \nadam
\end{aligned}
\end{equation*}
Note, some of these inclusions make use of the flexibility of hyperparameter schedules (dependence of $\psi^i$ on $t$). In particular, to approximate \momentum{} with \adam{}, one needs to choose a learning rate schedule that accounts for \adam{}'s bias correction.

If two optimizers have an inclusion relationship, the more general optimizer can never be worse with respect to \textit{any} metric of interest, provided the hyperparameters are sufficiently tuned to optimize that metric.
Optimally-tuned \momentum\ cannot underperform optimally-tuned \sgd, because setting $\gamma = 0$ in \momentum\ recovers \sgd. However, optimizers with more hyperparameters might be more expensive to tune, so we should have a theoretical or experimental reason for using (or creating) a more general optimizer. For example, $\momentum$ improves local convergence rates over $\sgd$ on twice-differentiable functions that are smooth and strongly convex \citep{polyak1964some}, and \nesterov\ has globally optimal convergence rates within the class of smooth and strongly convex functions \citep{nesterov1983method, nesterov2018lectures}.

At first glance, the taxonomy of optimizer inclusions appears to resolve many optimizer comparison questions. However, for a deep learning practitioner, there is no guarantee that the inclusion hierarchy is at all meaningful in practice. For example, the hyperparameters that allow \adam\ to match or outperform \momentum\ might not be easily accessible. They might exist only in the limit of very large values, or be so difficult to find that only practitioners with huge computational budgets can hope to discover them. Indeed, empirical studies and conventional wisdom hold that the inclusion hierarchy does not predict optimizer performance for many practical workloads \citep{wilson2017marginal,balles2017dissecting,schneider2019deepobs}. Either these experimental investigations are too limited or the taxonomy of this section is of limited practical interest and provides no guidance about which optimizer to use on a real workload. In the following section we attempt to answer this question experimentally, and show that these inclusion relationships are meaningful in practice.

\section{Experiments}
\label{sec:experiments}

\begin{table*}[t!]
\caption{Summary of workloads used in experiments.}
\label{table:workloads}
\centering
\begin{tabular}{@{}lllllll@{}}
\toprule
\textbf{Task} & \textbf{\begin{tabular}[c]{@{}l@{}}Evaluation metric\end{tabular}} & \textbf{Model} & \textbf{Dataset} & \textbf{\begin{tabular}[c]{@{}l@{}}Target error\end{tabular}} & \textbf{\begin{tabular}[c]{@{}l@{}}Batch size\end{tabular}} & \textbf{Budget} \\ \midrule
\multirow{5}{*}{\begin{tabular}[c]{@{}l@{}}Image \\ classification\end{tabular}} & \multirow{5}{*}{\begin{tabular}[c]{@{}l@{}}Classification \\ error\end{tabular}} & Simple CNN & Fashion MNIST & 6.6\% & 256 & 10k steps \\ \cmidrule(l){3-7} 
 &  & ResNet-32 & CIFAR-10 & 7\% & 256 & 50k steps \\ \cmidrule(l){3-7} 
 &  & CNN & CIFAR-100 & -- & 256 & 350 epochs \\ \cmidrule(l){3-7} 
 &  & VGG-16 & CIFAR-10 & -- & 128 & 250 epochs \\ \cmidrule(l){3-7} 
 &  & ResNet-50 & ImageNet & 24\% & 1024 & 150k steps \\ \midrule
\multirow{2}{*}{\begin{tabular}[c]{@{}l@{}}Language\\ modeling\end{tabular}} & \begin{tabular}[c]{@{}l@{}}Classification\\ error\end{tabular} & LSTM & War and Peace & -- & 50 & 200 epochs \\ \cmidrule(l){2-7} 
 & Cross entropy & Transformer & LM1B & 3.45 & 256 & 750k steps \\ \bottomrule
\end{tabular}
\end{table*}

An empirical comparison of optimizers should aim to inform a careful practitioner.
Accordingly, we model our protocol on a practitioner that is allowed to vary all optimization hyperparameters for each optimizer (e.g. $\alpha_t$, $\beta_1$, $\beta_2$, $\epsilon$ for \adam) in addition to a parameterized learning rate decay schedule, in contrast to studies that fix a subset of the optimization hyperparameters to their default values \citep[e.g.][]{wilson2017marginal,schneider2019deepobs}. There is no standard method for selecting the values of these hyperparameters, but most practitioners tune at least a subset of the optimization hyperparameters by running a set of trials to maximize performance over the validation set. In our experiments, we run tens to hundreds of individual trials per workload. Given the variety of workloads we consider, this trial budget covers a wide range of computational budgets. 

Selecting the hyperparameter search space for each optimizer is a key methodological choice for any empirical comparison of optimizers.  Prior studies have attempted to treat each optimizer fairly by using the ``same'' search space for all optimizers \citep[e.g.][]{wilson2017marginal,schneider2019deepobs}. However, this requires the assumption that similarly-named hyperparameters should take similar values between optimizers, which is not generally true. For example, \momentum\ and \nesterov\ both have similar-looking momentum and learning rate hyperparameters, but \nesterov\ tolerates larger values of its momentum \citep{sutskever2013importance}. The situation worsens with less closely related optimizers: similarly-named hyperparameters could have totally different units, making it impossible to equate search volumes. Despite coming with its own set of challenges, it is most informative to compare optimizers assuming the practitioner is allowed to tune hyperparameters for different optimizers independently by way of optimizer-specific search spaces.

In our experiments, we chose the search space for each optimizer by running an initial set of experiments over a relatively large search space. In a typical case, we ran a single set of initial trials per optimizer to select the final search space. However, in some cases we chose one of the search spaces poorly, so we ran another set of experiments to select the final search space.
We include these initial search spaces in Appendix~\ref{appendix:search-spaces} for the sake of transparency.
The effort required to choose each search space cannot easily be quantified: our initial guesses were inevitably informed by prior experience with particular models and optimizers.
This is true of all search spaces in the literature: tuning practices tend to be refined over many experiments and across many workloads, representing the sum total of our community's experience.

Following standard practice, we tuned the hyperparameters $\eta_0$, $1 - \gamma$, $1 - \beta_1$, and $1 - \beta_2$ from Table~\ref{table:updaterules} on a log scale by searching over the logarithms of their values. Additionally, we decoupled $\epsilon$ from the initial learning rate $\alpha_0$ by searching over the logarithms of $(\epsilon, \alpha_0 / \sqrt{\epsilon})$ for \rmsprop\ and   $(\epsilon, \alpha_0 / \epsilon)$ for \adam\ and \nadam, instead of $(\epsilon, \alpha_0)$. The fact that $\epsilon$ is coupled with the learning rate, with larger values of $\epsilon$ generally requiring larger learning rates, was shown for \adam\ by \citet{savarese2019domain}. %
These search space transforms merely served to make our hyperparameter search more efficient; in principle, our results would be the same if we used a larger trial budget and naively searched all hyperparameters on a linear scale.

We validated our final search spaces by checking that that the optimal hyperparameter values were away from the search space boundaries for all optimizers in all experiments (see Figure~\ref{fig:example_hparam_plot} in Appendix~\ref{appendix:extra-plots}). We provide our final search spaces for all experiments in Appendix~\ref{appendix:search-spaces}. The fact that our final error rates compare favorably to prior published results -- including reaching state-of-the-art for our particular configuration of ResNet-50 on ImageNet (see Section \ref{sec:our_first_results}) -- supports our claim that our methodology is highly competitive with expert tuning procedures. 

\subsection{Overview of Workloads and Experimental Details}

We investigated the relative performance of optimizers across a variety of image classification and language modeling tasks. For image classification, we trained a simple convolutional neural network (Simple CNN) on Fashion MNIST \citep{xiao2017fashion}; ResNet-32 \citep{he2016deep} on CIFAR-10 \citep{krizhevsky2009learning}; a CNN on CIFAR-100;
VGG-16 \citep{simonyan2014very} on CIFAR-10; and ResNet-50 on ImageNet \citep{russakovsky2015imagenet}. For language modeling, we trained a 2-layer LSTM model \citep{hochreiter1997long} on Tolstoy's \textit{War and Peace}; and Transformer \citep{vaswani2017attention} on LM1B \citep{chelba2014one}. We used a linear learning rate decay schedule parameterized the same way as \citet{shallue2019batch} for all workloads. We used a fixed batch size and a fixed budget of training steps for each workload independent of the optimizer. Table~\ref{table:workloads} summarizes these workloads and Appendix~\ref{appendix:workload-details} provides the full details.

\begin{figure*}[t!]
    \centering
    \includegraphics[width=\linewidth]{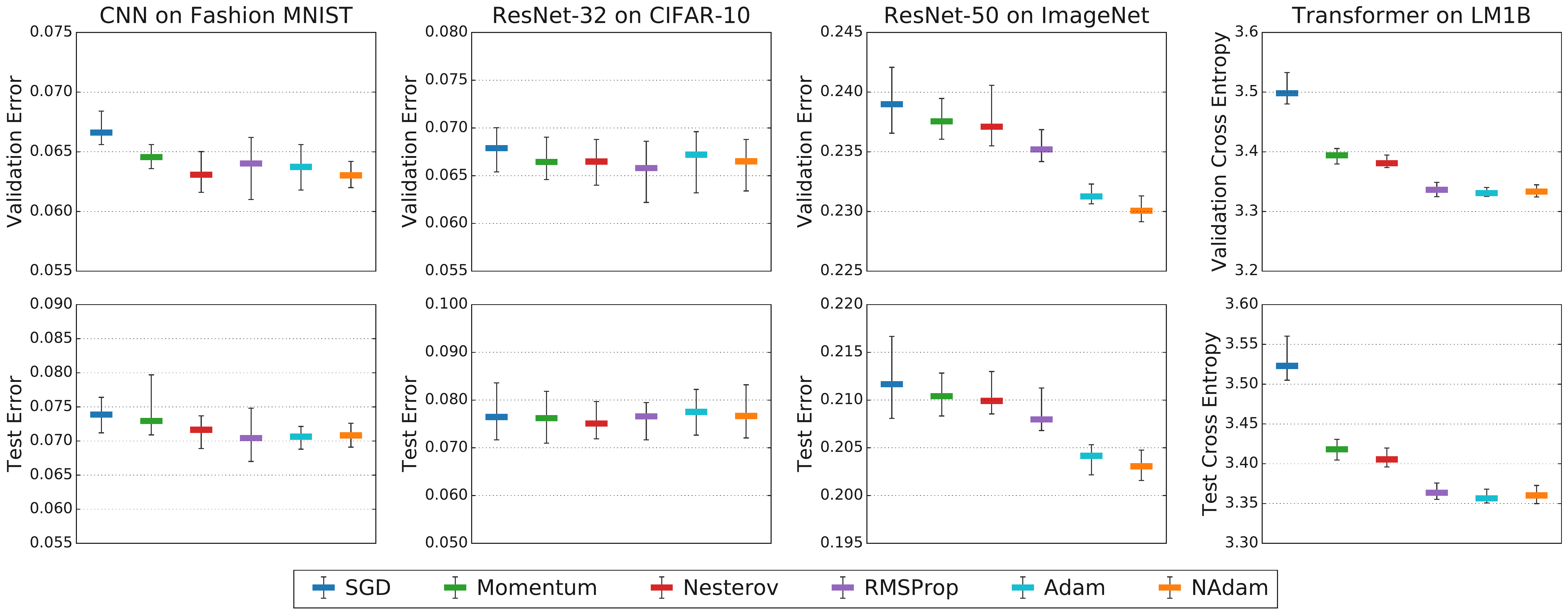}
    \caption{The relative performance of optimizers is consistent with the inclusion relationships, regardless of whether we compare final validation error (top) or test error (bottom). For all workloads, we tuned the hyperparameters of each optimizer separately, and selected the trial that achieved the lowest final validation error. Optimizers appear in the same order as the legend in all plots in this paper.}
    \label{fig:our_workloads_val_test}
\end{figure*}

\begin{figure*}[t!]
    \centering
    \includegraphics[width=\linewidth]{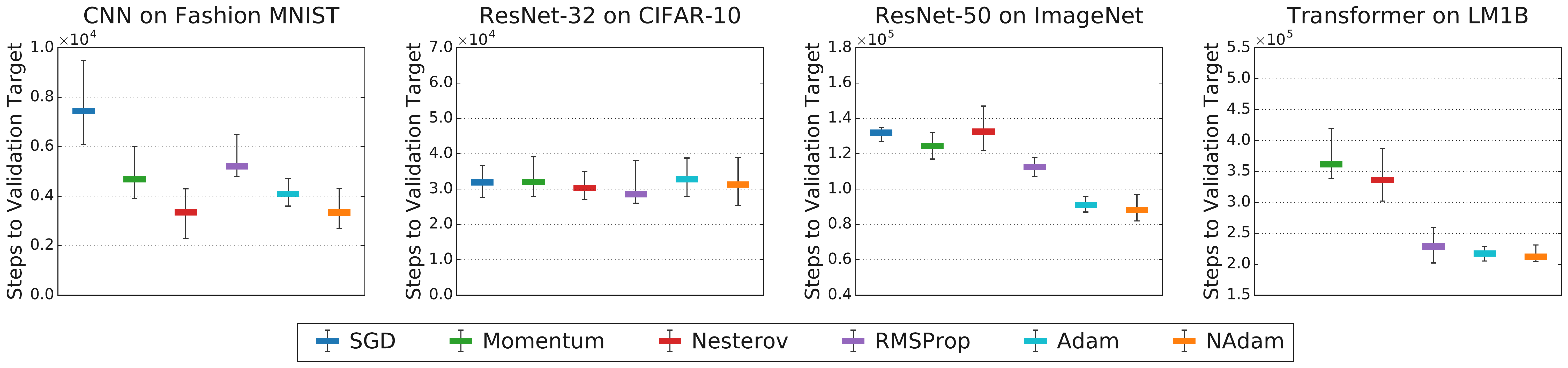}
    \caption{The relative training speed of optimizers is consistent with the inclusion relationships. We measured (idealized) training speed as the number of training steps required to reach a target validation error (see Table~\ref{table:workloads} for the error targets).}
    \label{fig:our_workloads_stt}
\end{figure*}

Given a search space, our tuning protocol sought to model a practitioner trying to achieve the best outcome with a fixed budget of trials ($10$, $50$, or $100$ depending on the workload).\footnote{In retrospect the best validation error across tuning trials converged quite quickly for our final search spaces, producing similar results with fewer than 20 trials in many cases. See Figures~\ref{fig:bootstrap-trial-lm1b-transformer}--~\ref{fig:bootstrap-trial-tolstoy-lstm} in Appendix~\ref{appendix:extra-plots}.} A feasible trial is any trial that achieves finite training loss. We used quasi-random uniform search \citep{BousquetEtAl_LDS_2017}, and continued the search until we obtained a fixed number of feasible trials. From those trials we considered two statistics. The first, in order to characterize the best outcome, is a metric of interest (e.g. test accuracy) corresponding to the trial achieving the optimum of some other metric (e.g. validation accuracy). The second, in order to characterize the speed of training, is the number of steps required to reach a fixed validation target conditional on at least one trial in the search having reached that target. We chose the target for each workload based on initial experiments and known values from the literature (see Table~\ref{table:workloads}). We estimated means and uncertainties using the bootstrap procedure described in Appendix~\ref{appendix:bootstrap}.

\subsection{Inclusion relationships matter in practice}
\label{sec:our_first_results}

Figure~\ref{fig:our_workloads_val_test} shows the final predictive performance of six optimizers on four different workloads after tuning hyperparameters to minimize validation error. Regardless of whether we compare final validation error or test error, the inclusion relationships hold in all cases -- a more general optimizer never underperforms any of its specializations within the error bars. Similar results hold for training error (see Figure~\ref{fig:our_workloads_train} in Appendix~\ref{appendix:extra-plots}). Training speed is also an important consideration, and Figure~\ref{fig:our_workloads_stt} demonstrates that the inclusion relationships also hold within error bars when we compare the number of steps required to reach a target validation error. Moreover, these results confirming the relevance of optimizer inclusion relationships do not depend on the exact step budgets or error targets we chose (see Figure~\ref{fig:budgets-and-targets} in Appendix~\ref{appendix:extra-plots}), although large changes to these values would require new experiments.

\begin{figure*}[t!]
    \centering
    \includegraphics[width=0.93\linewidth]{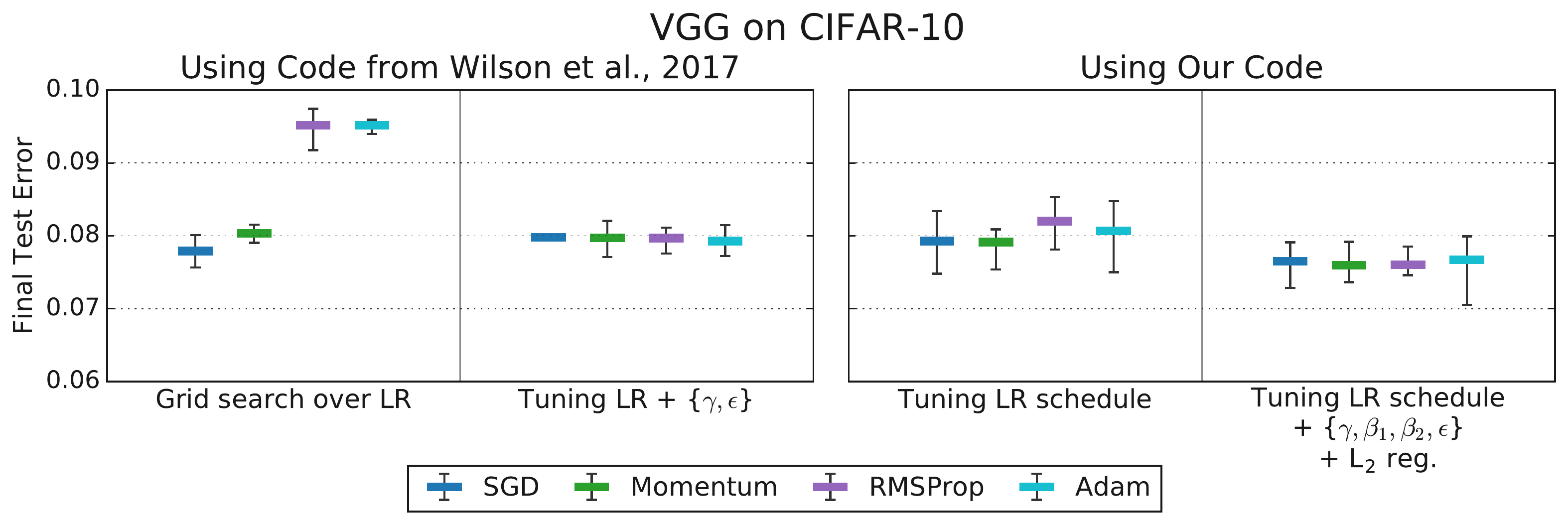}
    \caption{Tuning more hyperparameters removes the differences in test error between optimizers observed by \citet{wilson2017marginal}. Tuning a subset of optimizer hyperparameters and the initial learning rate is sufficient to equalize performance between all optimizers (left). More extensive hyperparameter tuning in our setup, including the learning rate schedule, improves results for all optimizers and still does not produce any differences between optimizer performances (right).}
    \label{fig:marginal_values}
\end{figure*}

Of course, just because a more general optimizer is no worse than any of its specializations doesn't mean the choice of optimizer makes a large difference on all workloads. For some workloads in Figures~\ref{fig:our_workloads_val_test} and~\ref{fig:our_workloads_stt}, all optimizers perform about the same, while other workloads have a clear ranking or even dramatic differences. For example, the choice of optimizer seems to make little difference for ResNet-32 on CIFAR-10; all optimizers achieve similar predictive performance and training speed.
On the other hand, Transformer on LM1B exhibits a clear ranking in terms of predictive performance and training speed. For this workload, $\adam$ needs only 60\% as many steps as $\momentum$ to reach our target error, and only 25\% as many steps to get the same final result as \sgd\ (see Figure~\ref{fig:budgets-and-targets} in the Appendix). These differences are clearly large enough to matter to a practitioner, and highlight the practical importance of choosing the right optimizer for some workloads.

The most general optimizers we considered were $\rmsprop$, $\adam$, and $\nadam$, which do not include each other as special cases, and whose relative performance is not predicted by inclusion relationships. Across the workloads we considered, none of these optimizers emerged as the clear winner, although \adam\ and \nadam\ generally seemed to have an edge over \rmsprop. For all of these optimizers, we sometimes had to set the $\epsilon$ parameter orders of magnitude larger than the default value in order to get good results. In particular, we achieved a validation accuracy of 77.1\% for ResNet-50 on ImageNet using \nadam\ with $\epsilon = 9475$, a result that exceeds the 76.5\% achieved by \citet{goyal2017accurate} using \momentum. Across just these 4 workloads, the range of the optimal values of the $\epsilon$ parameter spanned 10 orders of magnitude.

\subsection{Reconciling disagreements with previous work}
\label{sec:reconcilingdisagreements}

In order to confirm that differences in hyperparameter tuning protocols explain the differences between our conclusions and those of \citet{wilson2017marginal} and \citet{schneider2019deepobs}, we reproduced a representative subset of their results and then inverted, or at least collapsed, the ranking over optimizers just by expanding the hyperparameter search space.

\begin{figure*}[t!]
    \centering
    \includegraphics[width=0.93\linewidth]{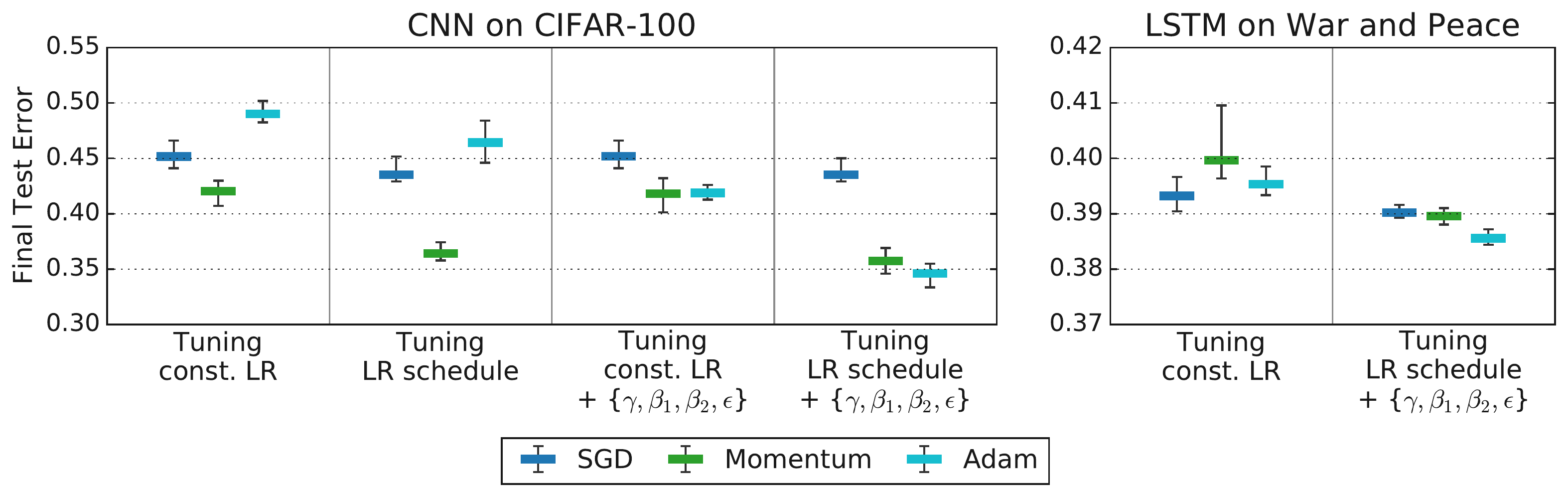}
    \caption{Tuning more hyperparameters changes optimizer rankings from \citet{schneider2019deepobs} to rankings that are consistent with the inclusion relationships. The leftmost columns for each workload reproduce the rankings from \citet{schneider2019deepobs}, while the remaining columns tune over increasingly general search spaces. All columns use our random search tuning protocol.}
    \label{fig:deepobs}
\end{figure*}

The left pane of Figure~\ref{fig:marginal_values} shows our experiments on VGG on CIFAR-10 using code released by \citet{wilson2017marginal}. When we match their protocol and perform their grid search over the initial learning rate and no other tuning, we reproduce their original result showing worse test error for \rmsprop{} and \adam{}. However, when we tune the momentum parameter and $\epsilon$ with random search, all four optimizers reach nearly identical test error rates.\footnote{\citet{wilson2017marginal} selected trials to minimize the training loss and then report test set results. As Figure~\ref{fig:marginal_values} shows, removing this somewhat non-standard choice and tuning on a validation set and reporting test set results does not change anything.} With our learning rate schedule search space, merely tuning the learning rate schedule was enough to make all optimizers reach the same test error within error bars. When we additionally tuned the optimization hyperparameters and weight decay in our setup we also get similar results for all optimizers, removing any evidence the inclusion relationships might be violated in practice.

Figure~\ref{fig:deepobs} shows our results with different tuning protocols for a CNN on CIFAR-100 and an LSTM language model trained on \textit{War and Peace} to match the experiments in \citet{schneider2019deepobs}. As reported by \citet{schneider2019deepobs}, if we only tune the learning rate without tuning the decay schedule or other optimizer hyperparameters, $\adam$ does worse than $\momentum$ for the CNN and $\sgd$ performs slightly better than $\adam$ and $\momentum$ on the \textit{War and Peace} dataset, although \citet{schneider2019deepobs} found a larger advantage for $\sgd$. However, once we tune the all the optimizer hyperparameters, $\adam$ does better than $\momentum$ which does better than $\sgd$, as predicted by the inclusion relationships.

We conclude that the reason both \citet{schneider2019deepobs} and \citet{wilson2017marginal} observed a ranking that, at first glance, contradicts the inclusion relationships is because they were not tuning enough of the hyperparameters. If we recast their results in our terminology where $\adam$ with default $\epsilon$ is a different optimizer than $\adam$ with $\epsilon$ tuned then there is no contradiction with our results and it becomes clear immediately that they do not consider the most interesting form of $\adam$ for practitioners.

\section{Conclusions}
\label{sec:conclusions}

Inspired by the recent efforts of \citet{wilson2017marginal} and \citet{schneider2019deepobs}, we set out to provide a detailed empirical characterization of the optimizer selection process in deep learning. Our central finding is that inclusion relationships between optimizers are meaningful in practice. When tuning all available hyperparameters under a realistic protocol at scales common in deep learning, we find that more general optimizers never underperform their special cases.
In particular, we found that \rmsprop{}, \adam{}, and \nadam{} never underperformed \sgd{}, \nesterov{}, or \momentum{} under our most exhaustive tuning protocol. We did not find consistent trends when comparing optimizers that could not approximate each other. We also found workloads for which there was not a statistically significant separation in the optimizer ranking.

Our experiments have some important limitations and we should be careful not to overgeneralize from our results. The first major caveat is that we did not measure the effects of varying the batch size. Recent empirical work \citep{shallue2019batch, zhang2019nqp} has shown that increasing the batch size can increase the gaps between training times for different optimizers, with the gap from \sgd{} to \momentum{} \citep{shallue2019batch} and from \momentum{}  to \adam{} \citep{zhang2019nqp} increasing with the batch size. Nevertheless, we strongly suspect that the inclusion relations would be predictive at any batch size under a tuning protocol similar to the one we used. The second important caveat of our results is that they inevitably depend on the tuning protocol and workloads that we considered. Although we made every attempt to conduct realistic experiments, we should only expect our detailed findings to hold for similar workloads under similar protocols, namely uniform quasi-random tuning for tens to hundreds of trials, over hypercube search spaces, and with our specific learning rate schedule parameterization. Nevertheless, these caveats reinforce our central point: all empirical comparisons of neural network optimizers depend heavily on the hyperparameter tuning protocol, perhaps far more than we are used to with comparisons between model architectures. 

If we were to extract ``best practices'' from our findings, then we suggest the following. If we can afford tens or more runs of our code, we should tune all of the hyperparameters of the popular adaptive gradient methods. Just because two hyperparameters have a similar role in two different update rules doesn't mean they should take similar values--- optimization hyperparameters tend to be coupled and the optimal value for one may depend on how the others are set. Our results also confirm that the optimal value of Adam's $\epsilon$ is problem-dependent, so the onus is on empirical studies that fix $\epsilon=10^{-8}$ to defend that choice. Finally, we should be skeptical of empirical comparisons of optimizers in papers, especially if an optimizer underperforms any of its specializations. When we do inevitably compare optimizers, we should report search spaces and highlight decisions about what hyperparameters were tuned when interpreting results.

\def\useAcks{0}

\if \useAcks 1
\subsubsection*{Acknowledgments}
We would like to thank Roy Frostig for his indispensable help throughout this project. We would also like to thank Alex Passos for help with NAdam implementations, Sam Smith for providing feedback on a draft, and Luke Metz for many helpful discussions. CJM is supported by the James D. Wolfensohn Fund.
\fi

\bibliography{refs}
\bibliographystyle{icml2020}

\newpage
\onecolumn
\appendix
\section{Optimizer inclusions}\label{appendix:inclusions}

Table~\ref{table:updaterules} summarizes the update rules for the optimizers we consider in this work. We assume update rules as implemented in TensorFlow r1.15. 
Here we prove their inclusion relationships, see Definition \ref{def:inclusion}.

\textbf{\momentum{} can exactly implement \sgd}

$\momentum(I_t, \eta_t, 0) = \sgd(I_t, \eta_t )$, so $\sgd \subseteq \momentum$.

\textbf{\nesterov{} can exactly implement \sgd}

$\nesterov(I_t, \eta_t, 0) = \sgd(I_t, \eta_t )$, so $\sgd \subseteq \nesterov$.

\textbf{\rmsprop{} with momentum can exactly implement \momentum}

Consider $\rmsprop(I_t, \eta_t, \gamma, \rho=1, \epsilon=0)$, so that
\begin{align*}
m_{t+1} &= \gamma m_{t} + \eta_t \nabla \loss(\theta_{t})\,,\\
\theta_{t+1} &= \theta_{t} - m_{t+1} \,.
\end{align*}
This is equivalent to \momentum{}, since
\begin{equation*}
    m_{t+1}^{\text{(\rmsprop)}} \equiv \eta_t v_{t+1}^{\text{(\momentum)}} \,.
\end{equation*}
Thus $\rmsprop(I_t, \eta_t, \gamma, 1, 0) = \momentum(I_t, \eta_t, \gamma)$, so $\momentum \subseteq \rmsprop$.

\textbf{\rmsterov{} can exactly implement \nesterov}

Consider $\rmsterov(I_t, \eta_t, \gamma, \rho=1, \epsilon=0)$, so that
\begin{align*}
m_{t+1} &= \gamma m_{t} + \eta_t \nabla \loss(\theta_{t})\,,\\
\theta_{t+1} &= \theta_{t} - \left[\gamma m_{t+1} + \eta_t \nabla \loss(\theta_{t}) \right]\,.
\end{align*}
This is equivalent to \momentum{}, since
\begin{equation*}
    m_{t+1}^{\text{(\rmsprop)}} \equiv \eta_t v_{t+1}^{\text{(\nesterov)}}\,.
\end{equation*}
Thus $\rmsterov(I_t, \eta_t, \gamma, 1, 0) = \momentum(I_t, \eta_t, \gamma)$, so $\momentum \subseteq \rmsterov$.

\textbf{\adam{} can approximate \momentum{} for large $\epsilon$}

Consider $\adam(I_t, \alpha_t=\epsilon \, \eta_t (1-\gamma^t), \beta_1 = \gamma, \beta_2 = 0, \epsilon)$, so that
\begin{align*}
    m_{t+1} &= \gamma m_{t} + (1 - \gamma) \nabla \loss(\theta_{t}) \,,&\\
    \theta_{t+1} &= \theta_{t} - \frac{\eta_t}{(1 - \gamma)} \left[ \frac{m_{t+1}}{|\nabla \loss(\theta_{t})| / \epsilon + 1} \right]\,.
\end{align*}
If $\epsilon$ is large, so that $|\nabla \loss(\theta_{t})| / \epsilon \ll 1$, then
\begin{align*}
    m_{t+1} &= \gamma m_{t} + (1 - \gamma) \nabla \loss(\theta_{t})\,, &\\
    \theta_{t+1} &= \theta_{t} - \eta_t \, \frac{m_{t+1}}{1-\gamma} \,.
\end{align*}
This is equivalent to \momentum{}, since \begin{equation*}
    m_t^{\text{(\adam)}} \equiv (1 - \gamma) \, v_t^{\text{(\momentum)}}.
\end{equation*}

Thus $\lim_{\epsilon \rightarrow \infty}\adam(I_t, \epsilon \, \eta_t (1-\gamma^t), \gamma, 0, \epsilon)= \momentum(I_t, \eta_t, \gamma)$, so $\momentum \subseteq \adam$

\textbf{\nadam{} can approximate \nesterov{} for large $\epsilon$}

Consider $\nadam(I_t, \alpha_t = \epsilon \, \eta_t (1-\gamma^t), \beta_1 = \gamma, \beta_2 = 0, \epsilon)$, so that
\begin{align*}
    m_{t+1} &= \gamma m_{t} + (1 - \gamma) \nabla \loss(\theta_{t}) \,,&\\
    \theta_{t+1} &= \theta_{t} - \frac{\eta_t}{(1 - \gamma)} \left[ \frac{\gamma m_{t+1} +(1-\gamma) \nabla \loss(\theta_{t})}{|\nabla \loss(\theta_{t})| / \epsilon + 1} \right] \,.
\end{align*}
If $\epsilon$ is large, so that $|\nabla \loss(\theta_{t})| / \epsilon \ll 1$, then
\begin{align*}
    m_{t+1} &= \gamma m_{t} + (1 - \gamma) \nabla \loss(\theta_{t})\,, &\\
    \theta_{t+1} &= \theta_{t} - \eta_t \, \left[\frac{\gamma m_{t+1}}{1-\gamma} + \nabla \loss(\theta_t) \right] \,.
\end{align*}
This is equivalent to \nesterov{}, since
\begin{equation*}
    m_t^{\text{(\nadam)}} \equiv (1 - \gamma) \, v_t^{\text{(\nesterov)}}
\end{equation*}
Thus $\lim_{\epsilon \rightarrow \infty}\nadam(I_t, \epsilon \, \eta_t (1-\gamma^t), \gamma, 0, \epsilon)= \nesterov(I_t, \eta_t, \gamma)$, so $\nesterov \subseteq \nadam$.

\section{Workload details}\label{appendix:workload-details}

This section details the datasets and models summarized in Table~\ref{table:workloads}.

\subsection{Dataset Descriptions}

For Fashion MNIST, CIFAR-10, ImageNet, and LM1B, our setup was identical to \citet{shallue2019batch} except for the image pre-processing details described below. For \textit{War and Peace}, our setup was identical to the ``Tolstoi'' dataset of \citet{schneider2019deepobs}.

\textbf{CIFAR-10/100:} We pre-processed images by subtracting the average value across all pixels and channels and dividing by the standard deviation.\footnote{We used the TensorFlow op \texttt{tf.image.per\_image\_standardization}.} For experiments with the ResNet-32 and CNN models, we followed the standard data augmentation scheme used in~\citet{he2016deep}: 4 pixels padded on each side with single random crop from padded image or its horizontal reflection. We did not use random cropping for experiments with VGG for consistency with \citet{wilson2017marginal}.

\textbf{ImageNet:} We augmented images at training time by resizing each image, taking a random crop of $224 \times 224$ pixels, randomly horizontally reflecting the cropped images, and randomly distorting the image colors. At evaluation time, we performed a single central crop of $224 \times 224$ pixels. In both training and evaluation, we then subtracted the global mean RGB value from each pixel using the values computed by \citet{simonyan2014very}.\footnote{See \url{https://gist.github.com/ksimonyan/211839e770f7b538e2d8\#description} for the mean RGB values used.}

\subsection{Model Descriptions}

\textbf{Simple CNN} is identical to the base model described in~\citet{shallue2019batch}. It consists of 2 convolutional layers with max pooling followed by 1 fully connected layer. The convolutional layers use $5 \times 5$ filters with stride 1, ``same" padding, and ReLU activation function. Max pooling uses a $2\times2$ window with stride 2. Convolutional layers have 32 and 64 filters each and the fully connected layer has 1024 units. It does not use batch normalization.

\textbf{CNN} is the ``All-CNN-C'' model from~\citet{springenberg2014striving}, as used in  \citet{schneider2019deepobs}.  The model consists of 3 convolutional layer blocks with max pooling. The convolutional layers use $5 \times 5$ filters with stride 1, ``same" padding, and ReLU activation function. Max pooling uses a $2\times2$ window with stride 2. Convolutional layer blocks have 96, 192 and 192 filters each. As in \citet{schneider2019deepobs}, we used $L_2$ regularization of \num{5e-4}. 

\textbf{ResNet} is described in~\citet{he2016deep}. We used the improved residual block described in~\citet{he2016identity}. We used batch normalization  \citep{ioffe2015batch} with exponential moving average (EMA) decay of 0.997 for ResNet-32, and ghost batch normalization~\citep{hoffer2017train} with ghost batch size of 32 and EMA decay of $0.9$ for ResNet-50.

\textbf{VGG} is based on ``model C'' from \citet{simonyan2014very}. It consists of 13 convolutional layers followed by 3 fully connected hidden layers. We followed the modification used by~\citet{wilson2017marginal} with batch normalization layers.

\textbf{LSTM} is a two hidden-layer LSTM model \citep{hochreiter1997long} identical to the model used in~\citet{schneider2019deepobs}. It uses 128 embedding dimensions and 128 hidden units.

\textbf{Transformer} is the ``base'' model described in \citep{vaswani2017attention}. We used it as an autoregressive language model by applying the decoder directly to the sequence of word embeddings for each sentence. Unlike the default implementation, we removed dropout regularization and used separate weight matrices for the input embedding layer and the pre-softmax linear transformation, as we observed these choices led to better performing models.

\section{Estimating trial outcomes via bootstrap}\label{appendix:bootstrap}

Our tuning protocol corresponds to running trials with quasi-random hyperparameter values sampled uniformly from the search space until $K$ feasible trials are obtained, with $K$ depending on the workload. We then select the best trial, based on our statistic of interest, over those $K$ trials.

We used the following bootstrap procedure to estimate means and uncertainties of our tuning protocol. We ran $N>K$ trials, with $N$ depending on the workload. Then, for each bootstrap sample, we resampled the dataset of $N$ trials with replacement and computed our statistic on the first $K$ trials of the resampled dataset. We collected $100$ such bootstrap samples each time, and from those computed the means, $5^\text{th}$ percentiles, and $95^\text{th}$ percentiles of the bootstrap distribution. We used this procedure to generate the means and error bars for each plot.

Simple CNN on Fashion MNIST used $(K,N)=(100, 500)$; ResNet-32 on CIFAR-10 used $(K,N)=(100, 500)$; ResNet-50 on ImageNet used $(K,N)=(50,250)$; Transformer on LM1B used $(K,N)=(50,250)$; VGG on CIFAR-10 with our code used $(K,N)=(50,250)$ for tuning the learning rate schedule and $(K,N)=(100, 500)$ for tuning the learning rate schedule, $\{\gamma, \beta_1, \beta_2, \epsilon\}$, and $L_2$ regularization; CNN on CIFAR-100 used $(K,N)=(100, 500)$; LSTM on \textit{War and Peace} used $(K,N)=(10,50)$ for tuning just the learning rate and $(K,N)=(100, 500)$ for tuning the learning rate schedule and $\{\gamma, \beta_1, \beta_2, \epsilon\}$.

The sole exceptions to this bootstrap procedure are the two left panels of Figure~\ref{fig:marginal_values}, for which we used a similar procedure to \citet{wilson2017marginal} to ensure comparability. For each optimizer, we selected the trial that minimized validation error in our final search space and ran the same hyperparameter values 5 times, reporting the mean, minimum, and maximum test error over those 5 runs in Figure~\ref{fig:marginal_values}. This is slightly different to \citet{wilson2017marginal}, who chose the trial that minimized training error and reported validation error. When tuning the learning rate and $\{ \gamma, \epsilon \}$, we used 24 trials per optimizer in the initial search space (which we used to select the final search space), and 16 trials per optimizer in the final search space.

\section{Hyperparameter Search Spaces}\label{appendix:search-spaces}

Below we report the search spaces used for our experiments. We include both the initial search spaces used to refine the search spaces, and the final spaces used to generate the plots. When only one search space was used, we denote the initial space as final. $\eta_0$, $\alpha_0$, $1 - \gamma$, $1 - \beta_1$, $1 - \beta_2$, $\epsilon$, and combinations thereof are always tuned on a log scale. The number of samples from each search space is specified in Appendix~\ref{appendix:bootstrap}.

\newcommand{\srange}[2] {\lbrack\num{#1}, \num{#2}\rbrack}
\newcommand{\specialcell}[2][c]{%
  \begin{tabular}[#1]{@{}c@{}}#2\end{tabular}}

\clearpage

\subsection{CNN on Fashion MNIST}

We used linear learning rate decay for all experiments. We tuned the number of decay steps within $\lbrack0.5, 1.0\rbrack$ times the number of training steps and the learning rate decay factor within $\{\num{e-3}, \num{e-2}, \num{e-1}\}$. We did not use $L_2$ regularization or weight decay.

\begin{table}[!htb]
\centering
\setlength{\extrarowheight}{3.5pt}
\begin{tabular}{|c|c|}
\hline
 & $\eta_0$\\ \hline
initial & $\srange{e-2}{e2}$ \\ \hline
final & $\srange{e-2}{e0}$ \\ \hline
\end{tabular}
\caption{\sgd}
\end{table}

\begin{table}[!htb]
\centering
\setlength{\extrarowheight}{3.5pt}
\begin{tabular}{|c|c|c|}
\hline
 & $\eta_0$ & $1 - \gamma$ \\ \hline
initial & $\srange{e-4}{e2}$ & $\srange{e-4}{e0}$ \\ \hline
final & $\srange{e-3}{e-1}$ & $\srange{e-3}{e0}$ \\ \hline
\end{tabular}
\caption{\momentum}
\end{table}

\begin{table}[!htb]
\centering
\setlength{\extrarowheight}{3.5pt}
\begin{tabular}{|c|c|c|}
\hline
 & $\eta_0$ & $1 - \gamma$ \\ \hline
initial & $\srange{e-4}{e2}$ & $\srange{e-4}{e0}$ \\ \hline
final & $\srange{e-3}{e-1}$ & $\srange{e-3}{e0}$ \\ \hline
\end{tabular}
\caption{\nesterov}
\end{table}

\begin{table}[!htb]
\centering
\setlength{\extrarowheight}{3.5pt}
\begin{tabular}{|c|c|c|c|c|}
\hline
 & $\eta_0 / \sqrt{\epsilon}$ & $1 - \gamma$ & $1 - \rho$ & $\epsilon$ \\ \hline
initial & $\srange{e-2}{e4}$ & $\srange{e-3}{e0}$ & $\srange{e-4}{e0}$ & $\srange{e-10}{e10}$ \\ \hline
final & $\srange{e-2}{e0}$ & $\srange{e-3}{e0}$ & $\srange{e-4}{e0}$ & $\srange{e-10}{e-6}$ \\ \hline
\end{tabular}
\caption{\rmsprop}
\end{table}

\begin{table}[!htb]
\centering
\setlength{\extrarowheight}{3.5pt}
\begin{tabular}{|c|c|c|c|c|}
\hline
 & $\alpha_0 / \epsilon$ & $1 - \beta_1$ & $1 - \beta_2$ & $\epsilon$ \\ \hline
initial & $\srange{e-2}{e-4}$ & $\srange{e-3}{e0}$ & $\srange{e-4}{e0}$ & $\srange{e-10}{e10}$ \\ \hline
final & $\srange{e-1}{e1}$ & $\srange{e-3}{e0}$ & $\srange{e-4}{e0}$ & $\srange{e-6}{e-2}$ \\ \hline
\end{tabular}
\caption{\adam}
\end{table}

\begin{table}[!htb]
\centering
\setlength{\extrarowheight}{3.5pt}
\begin{tabular}{|c|c|c|c|c|}
\hline
 & $\alpha_0 / \epsilon$ & $1 - \beta_1$ & $1 - \beta_2$ & $\epsilon$ \\ \hline
initial & $\srange{e-2}{e4}$ & $\srange{e-3}{e0}$ & $\srange{e-4}{e0}$ & $\srange{e-10}{e10}$ \\ \hline
final & $\srange{e-1}{e1}$ & $\srange{e-3}{e0}$ & $\srange{e-4}{e0}$ & $\srange{e-6}{e-2}$ \\ \hline
\end{tabular}
\caption{\nadam}
\end{table}

\clearpage

\subsection{ResNet-32 on CIFAR-10}
\label{sec:search_space_resnet_cifar}

We used linear learning rate decay for all experiments. We tuned the number of decay steps within $\lbrack0.5, 1.0\rbrack$ times the number of training steps and the learning rate decay factor $f$ within the values shown in the tables below. $\lambda_{L_2}$ denotes the $L_2$ regularization coefficient.

\begin{table}[!htb]
\centering
\setlength{\extrarowheight}{3.5pt}
\begin{tabular}{|c|c|c|c|}
\hline
 & $\eta_0$ & $\lambda_{L_2}$ & $f$ \\ \hline
initial & $\srange{e-2}{e2}$ & $\{\num{e-5}, \num{e-4}, \num{e-3}, \num{e-2}\}$ & $\{\num{e-4}, \num{e-3}, \num{e-2}, \num{e-1}\}$ \\ \hline
final & $\srange{e-1}{e1}$ & $\{\num{e-5}, \num{e-4}, \num{e-3}, \num{e-2}\}$ & $\{\num{e-4}, \num{e-3}, \num{e-2}, \num{e-1}\}$ \\ \hline
\end{tabular}
\caption{\sgd}
\end{table}

\begin{table}[!htb]
\centering
\setlength{\extrarowheight}{3.5pt}
\begin{tabular}{|c|c|c|c|c|}
\hline
 & $\eta_0$ & $1 - \gamma$ & $\lambda_{L_2}$ & $f$ \\ \hline
initial & $\srange{e-4}{e2}$ & $\srange{e-3}{e0}$ & $\{\num{e-5}, \num{e-4}, \num{e-3}, \num{e-2}\}$ & $\{\num{e-4}, \num{e-3}, \num{e-2}, \num{e-1}\}$ \\ \hline
final & $\srange{e-2}{e0}$ & $\srange{e-3}{e0}$ & $\{\num{e-5}, \num{e-4}, \num{e-3}, \num{e-2}\}$ & $\{\num{e-4}, \num{e-3}, \num{e-2}, \num{e-1}\}$ \\ \hline
\end{tabular}
\caption{\momentum}
\end{table}

\begin{table}[!htb]
\centering
\setlength{\extrarowheight}{3.5pt}
\begin{tabular}{|c|c|c|c|c|}
\hline
 & $\eta_0$ & $1 - \gamma$ & $\lambda_{L_2}$ & $f$ \\ \hline
initial & $\srange{e-4}{e2}$ & $\srange{e-4}{e1}$ & $\num{e-4}$ & $\{\num{e-3}, \num{e-2}, \num{e-1}\}$ \\ \hline
final & $\srange{e-2}{e0}$ & $\srange{e-3}{e0}$ & $\{\num{e-5}, \num{e-4}, \num{e-3}, \num{e-2}\}$ & $\{\num{e-4}, \num{e-3}, \num{e-2}, \num{e-1}\}$ \\ \hline
\end{tabular}
\caption{\nesterov}
\end{table}

\begin{table}[!htb]
\centering
\setlength{\extrarowheight}{3.5pt}
\begin{tabular}{|c|c|c|c|c|c|c|}
\hline
 & $\eta_0 / \sqrt{\epsilon}$ & $1 - \gamma$ & $1 - \rho$ & $\epsilon$ & $\lambda_{L_2}$ & $f$ \\ \hline
initial & $\srange{e-2}{e4}$ & $\srange{e-3}{e0}$ & $\srange{e-4}{e0}$ & $\srange{e-10}{e10}$ & $\num{e-4}$ & $\{\num{e-3}, \num{e-2}, \num{e-1}\}$ \\ \hline
final & $\srange{e-2}{e0}$ & $\srange{e-3}{e0}$ & $\srange{e-4}{e0}$ & $\srange{e-4}{e0}$ & \specialcell{$\{\num{e-5}, \num{e-4}$ \\ $\num{e-3}, \num{e-2}\}$} & \specialcell{$\{\num{e-4}, \num{e-3}$ \\ $\num{e-2}, \num{e-1}\}$} \\ \hline
\end{tabular}
\caption{\rmsprop}
\end{table}

\begin{table}[!htb]
\centering
\setlength{\extrarowheight}{3.5pt}
\begin{tabular}{|c|c|c|c|c|c|c|}
\hline
 & $\alpha_0 / \epsilon$ & $1 - \beta_1$ & $1 - \beta_2$ & $\epsilon$ & $\lambda_{L_2}$ & $f$ \\ \hline
initial & $\srange{e-4}{e-1}$ & $\srange{e-3}{5e-1}$ & $\srange{e-4}{e-1}$ & $\srange{e-9}{e-5}$ & $\num{e-4}$ & \specialcell{$\{\num{e-3}, \num{e-2}$ \\ $\num{e-1}\}$} \\ \hline
final & $\srange{e-1}{e1}$ & $\srange{e-3}{e0}$ & $\srange{e-4}{e0}$ & $\srange{e-3}{e1}$ & \specialcell{$\{\num{e-5}, \num{e-4}$ \\ $\num{e-3}, \num{e-2}\}$} & \specialcell{$\{\num{e-4}, \num{e-3}$ \\ $\num{e-2}, \num{e-1}\}$} \\ \hline
\end{tabular}
\caption{\adam}
\end{table}

\begin{table}[!htb]
\centering
\setlength{\extrarowheight}{3.5pt}
\begin{tabular}{|c|c|c|c|c|c|c|}
\hline
 & $\alpha_0 / \epsilon$ & $1 - \beta_1$ & $1 - \beta_2$ & $\epsilon$ & $\lambda_{L_2}$ & $f$ \\ \hline
initial & $\srange{e-2}{e4}$ & $\srange{e-3}{e0}$ & $\srange{e-4}{e0}$ & $\srange{e-10}{e10}$ & \specialcell{$\{\num{e-5}, \num{e-4}$ \\ $\num{e-3}, \num{e-2}\}$} & \specialcell{$\{\num{e-4}, \num{e-3}$ \\ $\num{e-2}, \num{e-1}\}$} \\ \hline
final & $\srange{e-2}{e0}$ & $\srange{e-3}{e0}$ & $\srange{e-4}{e0}$ & $\srange{e0}{e4}$ & \specialcell{$\{\num{e-5}, \num{e-4}$ \\ $\num{e-3}, \num{e-2}\}$} & \specialcell{$\{\num{e-4}, \num{e-3}$ \\ $\num{e-2}, \num{e-1}\}$} \\ \hline
\end{tabular}
\caption{\nadam}
\end{table}

\clearpage

\subsection{ResNet-50 on ImageNet}
\label{sec:search_space_resnet_imagenet}

We used linear learning rate decay for all experiments. We tuned the number of decay steps within $\lbrack0.5, 1.0\rbrack$ times the number of training steps and the learning rate decay factor $f$ within the values shown in the tables below. $\lambda_\text{wd}$ denotes the weight decay coefficient and $\tau$ denotes the label smoothing coefficient.

\begin{table}[!ht]
\centering
\setlength{\extrarowheight}{3.5pt}
\begin{tabular}{|c|c|c|c|c|}
\hline
 & $\eta_0$ & $\lambda_\text{wd}$ & $\tau$ & $f$ \\ \hline
initial & $\srange{e-2}{e1}$ & $\srange{e-5}{e-2}$ & $\{0, \num{e-2}, \num{e-1}\}$ & $\{\num{e-4}, \num{e-3}, \num{e-2}, \num{e-1}\}$ \\ \hline
final & $\srange{e0}{e2}$ & $\srange{e-4}{e-3}$ & $\num{e-1}$ & $\{\num{e-4}, \num{e-3}, \num{e-2}, \num{e-1}\}$ \\ \hline
\end{tabular}
\caption{\sgd}
\end{table}

\begin{table}[!htb]
\centering
\setlength{\extrarowheight}{3.5pt}
\begin{tabular}{|c|c|c|c|c|c|}
\hline
 & $\eta_0$ & $1 - \gamma$ & $\lambda_\text{wd}$ & $\tau$ & $f$ \\ \hline
initial & $\srange{e-3}{e0}$ & $\srange{e-3}{e0}$ & $\srange{e-5}{e-2}$ & $\{0, \num{e-2}, \num{e-1}\}$ & $\{\num{e-4}, \num{e-3}, \num{e-2}, \num{e-1}\}$ \\ \hline
final & $\srange{e-2}{e0}$ & $\srange{e-2}{e0}$ & $\srange{e-4}{e-3}$ & $\num{e-2}$ & $\{\num{e-4}, \num{e-3}, \num{e-2}, \num{e-1}\}$ \\ \hline
\end{tabular}
\caption{\momentum}
\end{table}

\begin{table}[!htb]
\centering
\setlength{\extrarowheight}{3.5pt}
\begin{tabular}{|c|c|c|c|c|c|}
\hline
 & $\eta_0$ & $1 - \gamma$ & $\lambda_\text{wd}$ & $\tau$ & $f$ \\ \hline
initial & $\srange{e-3}{e0}$ & $\srange{e-3}{e0}$ & $\srange{e-5}{e-2}$ & $\{0, \num{e-2}, \num{e-1}\}$ & $\num{e-3}$ \\ \hline
final & $\srange{e-2}{e0}$ & $\srange{e-3}{e0}$ & $\srange{e-4}{e-3}$ & $0$ & $\{\num{e-4}, \num{e-3}, \num{e-2}, \num{e-1}\}$ \\ \hline
\end{tabular}
\caption{\nesterov}
\end{table}

\begin{table}[!htb]
\centering
\setlength{\extrarowheight}{3.5pt}
\begin{tabular}{|c|c|c|c|c|c|c|c|}
\hline
 & $\eta_0 / \sqrt{\epsilon}$ & $1 - \gamma$ & $1 - \rho$ & $\epsilon$ & $\lambda_\text{wd}$ & $\tau$ & $f$ \\ \hline
initial & $\srange{e-2}{e4}$ & $0.1$ & $\srange{e-4}{e0}$ & $\srange{e-10}{e10}$ & $\srange{e-5}{e-2}$ & $\{0, \num{e-2}, \num{e-1}\}$ & $\num{e-3}$ \\ \hline
final & $\srange{e-2}{e0}$ & $0.1$ & $\srange{e-2}{e0}$ & $\srange{e-8}{e-3}$ & $\srange{e-4}{e-3}$ & $0$ & \specialcell{$\{\num{e-4}, \num{e-3}$ \\ $\num{e-2}, \num{e-1}\}$} \\ \hline
\end{tabular}
\caption{\rmsprop}
\end{table}

\begin{table}[!htb]
\centering
\setlength{\extrarowheight}{3.5pt}
\begin{tabular}{|c|c|c|c|c|c|c|}
\hline
 & $\alpha_0 / \epsilon$ & $1 - \beta_1$ & $\epsilon$ & $\lambda_\text{wd}$ & $\tau$ & $f$ \\ \hline
initial & $\srange{e0}{e2}$ & $\srange{e-3}{e0}$ & $\srange{e0}{e4}$ & $\srange{e-5}{e-3}$ & $\{0, \num{e-2}, \num{e-1}\}$ & $\num{e-3}$ \\ \hline
final & $\srange{e0}{e2}$ & $\srange{e-2}{e0}$ & $\srange{e-2}{e2}$ & $\num{e-4}$ & $\num{e-1}$ & \specialcell{$\{\num{e-4}, \num{e-3}$ \\ $\num{e-2}, \num{e-1}\}$} \\ \hline
\end{tabular}
\caption{\adam}
\end{table}
 
\begin{table}[!htb]
\centering
\setlength{\extrarowheight}{3.5pt}
\begin{tabular}{|c|c|c|c|c|c|c|}
\hline
 & $\alpha_0 / \epsilon$ & $1 - \beta_1$ & $\epsilon$ & $\lambda_\text{wd}$ & $\tau$ & $f$ \\ \hline
initial & $\srange{e-1}{e3}$ & $\srange{e-3}{e0}$ & $\srange{e-2}{e10}$ & $\srange{e-5}{e-2}$ & $\{0, \num{e-2}, \num{e-1}\}$ & $\num{e-3}$ \\ \hline
final & $\srange{e0}{e2}$ & $\srange{e-3}{e0}$ & $\srange{e3}{e7}$ & $\num{e-4}$ & $\num{e-1}$ & $\num{e-3}$ \\ \hline
\end{tabular}
\caption{\nadam}
\end{table}

\clearpage

\subsection{Transformer on LM1B}
\label{sec:search_space_transformer_lm1b}
We used linear learning rate decay for all experiments. We tuned the number of decay steps within $\lbrack0.5, 1.0\rbrack$ times the number of training steps and the learning rate decay factor within $\{\num{e-4}, \num{e-3}, \num{e-2}, \num{e-1}, \num{e0}\}$.

\begin{table}[!htb]
\centering
\setlength{\extrarowheight}{3.5pt}
\begin{tabular}{|c|c|}
\hline
 & $\eta_0$\\ \hline
initial & $\srange{e-4}{e-1}$ \\ \hline
final & $\srange{e-3}{e-1}$ \\ \hline
\end{tabular}
\caption{\sgd}
\end{table}

\begin{table}[!htb]
\centering
\setlength{\extrarowheight}{3.5pt}
\begin{tabular}{|c|c|c|}
\hline
initial & $\srange{e-4}{e-1}$ & $\srange{e-4}{e0}$ \\ \hline
final & $\srange{e-4}{e-2}$ & $\srange{e-3}{e0}$ \\ \hline
\end{tabular}
\caption{\momentum}
\end{table}

\begin{table}[!htb]
\centering
\setlength{\extrarowheight}{3.5pt}
\begin{tabular}{|c|c|c|}
\hline
 & $\eta_0$ & $1 - \gamma$ \\ \hline
initial & $\srange{e-4}{e-1}$ & $\srange{e-4}{e0}$ \\ \hline
final & $\srange{e-4}{e-2}$ & $\srange{e-3}{e0}$ \\ \hline
\end{tabular}
\caption{\nesterov}
\end{table}

\begin{table}[!htb]
\centering
\setlength{\extrarowheight}{3.5pt}
\begin{tabular}{|c|c|c|c|c|}
\hline
 & $\eta_0 / \sqrt{\epsilon}$ & $1 - \gamma$ & $1 - \rho$ & $\epsilon$ \\ \hline
initial & $\srange{e-2}{e4}$ & $\srange{e-3}{e0}$ & $\srange{e-4}{e0}$ & $\srange{e-10}{e10}$ \\ \hline
final & $\srange{e-1}{e1}$ & $\srange{e-3}{e0}$ & $\srange{e-4}{e0}$ & $\srange{e-9}{e-5}$ \\ \hline
\end{tabular}
\caption{\rmsprop}
\end{table}

\begin{table}[!htb]
\centering
\setlength{\extrarowheight}{3.5pt}
\begin{tabular}{|c|c|c|c|c|}
\hline
 & $\alpha_0 / \epsilon$ & $1 - \beta_1$ & $1 - \beta_2$ & $\epsilon$ \\ \hline
initial & $\srange{e-2}{e4}$ & $\srange{e-3}{e0}$ & $\srange{e-4}{e0}$ & $\srange{e-10}{e10}$ \\ \hline
final & $\srange{e1}{e3}$ & $\srange{e-3}{e0}$ & $\srange{e-6}{e-2}$ & $\srange{e-7}{e-3}$ \\ \hline
\end{tabular}
\caption{\adam}
\end{table}

\begin{table}[!htb]
\centering
\setlength{\extrarowheight}{3.5pt}
\begin{tabular}{|c|c|c|c|c|}
\hline
 & $\alpha_0 / \epsilon$ & $1 - \beta_1$ & $1 - \beta_2$ & $\epsilon$ \\ \hline
initial & $\srange{e-2}{e4}$ & $\srange{e-3}{e0}$ & $\srange{e-4}{e0}$ & $\srange{e-10}{e10}$ \\ \hline
final & $\srange{e0}{e2}$ & $\srange{e-3}{e0}$ & $\srange{e-6}{e-2}$ & $\srange{e-6}{e-2}$ \\ \hline
\end{tabular}
\caption{\nadam}
\end{table}

\clearpage

\subsection{VGG on CIFAR-10 Using Code from \texorpdfstring{\citet{wilson2017marginal}}{Marginal Values}}
\subsubsection{Grid Search over Learning Rate}
We tuned over the same grid of initial learning rate values for each optimizer as \citet{wilson2017marginal}. As in \citet{wilson2017marginal}, we decayed the initial learning rate by a factor of $0.5$ every 25 epochs and used a fixed $L_2$ regularization coefficient of $0.0005$. 

\subsubsection{Tuning Learning Rate \& \texorpdfstring{$\{\gamma, \epsilon\}$}{Optimizer hyperparameters}}

We used our quasi-random tuning protocol to tune over the initial learning rate, $\momentum$'s $\gamma$, $\rmsprop$'s $\epsilon$, and $\adam$'s $\epsilon$. As in \citet{wilson2017marginal}, we decayed the initial learning rate by a factor of $0.5$ every 25 epochs and used a fixed $L_2$ regularization coefficient of $0.0005$.

\begin{table}[!htb]
\centering
\setlength{\extrarowheight}{3.5pt}
\begin{tabular}{|c|c|}
\hline
 & $\eta_0$\\ \hline
initial & $\srange{e-3}{e0}$ \\ \hline
final & $\srange{e-1}{e1}$ \\ \hline
\end{tabular}
\caption{\sgd}
\end{table}

\begin{table}[!htb]
\centering
\setlength{\extrarowheight}{3.5pt}
\begin{tabular}{|c|c|c|}
\hline
 & $\eta_0$ & $1 - \gamma$ \\ \hline
initial & $\srange{e-3}{e0}$ & $\srange{e-3}{e0}$ \\ \hline
final & $\srange{e-1}{e1}$ & $\srange{e-1}{e0}$ \\ \hline
\end{tabular}
\caption{\momentum}
\end{table}

\begin{table}[!htb]
\centering
\setlength{\extrarowheight}{3.5pt}
\begin{tabular}{|c|c|c|}
\hline
 & $\epsilon$ & ${\alpha_0}/{\sqrt{\epsilon}}$ \\ \hline
initial & $\srange{e-10}{e10}$ & $\srange{e-2}{e4}$ \\ \hline
final & $\srange{e-2}{e2}$ & $\srange{e-1}{e1}$ \\ \hline
\end{tabular}
\caption{\rmsprop}
\end{table}

\begin{table}[!htb]
\centering
\setlength{\extrarowheight}{3.5pt}
\begin{tabular}{|c|c|c|}
\hline
 & $\epsilon$ & $\alpha_0 / \epsilon$ \\ \hline
initial & $\srange{e-10}{e10}$ & $\srange{e-2}{e4}$ \\ \hline
final & $\srange{e6}{e10}$ & $\srange{e-1}{e1}$ \\ \hline
\end{tabular}
\caption{\adam}
\end{table}

\clearpage

\subsection{VGG on CIFAR-10 Using our Code}

We used linear learning rate decay for all experiments. We tuned the number of decay steps within $\lbrack0.5, 1.0\rbrack$ times the number of training steps and the learning rate decay factor within $\{\num{e-4}, \num{e-3}, \num{e-2}, \num{e-1}\}$.

\subsubsection{Tuning Learning Rate Schedule}
We fixed all optimizer hyperparameters excluding the learning rate to match those specified in \citet{wilson2017marginal}. As in \citet{wilson2017marginal}, we used a fixed $L_2$ regularization coefficient of $0.0005$.

\begin{table}[!htb]
\centering
\setlength{\extrarowheight}{3.5pt}
\begin{tabular}{|c|c|c|c|c|}
\hline
& $\eta_0$ (\sgd) & $\eta_0$ (\momentum) & $\eta_0$ (\rmsprop) & $\alpha_0$ (\adam) \\ \hline
initial & $\srange{e-3}{e1}$ & $\srange{e-3}{e1}$ & $\srange{e-5}{e-1}$ & $\srange{e-5}{e-1}$\\ \hline
final & 1.0 & $\srange{e-2}{e0}$ & $\srange{e-4}{e-2}$ & $\srange{e-5}{e-1}$\\ \hline
\end{tabular}
\caption{Learning rate search ranges.}
\end{table}

\subsubsection{Tuning Learning Rate Schedule \& \texorpdfstring{$\{\gamma, \beta_1, \beta_2, \epsilon, \lambda_{L_2}\}$}{Optimizer hyperparameters \& Weight Decay}}

\begin{table}[!htb]
\centering
\setlength{\extrarowheight}{3.5pt}
\begin{tabular}{|c|c|c|}
\hline
 & $\eta_0$ & $\lambda_{L_2}$ \\ \hline
initial & $\srange{e-3}{e1}$ & $\srange{e-5}{e-2}$ \\ \hline
final & $\srange{e-2}{e0}$ & $\srange{e-3}{e-1}$ \\ \hline
\end{tabular}
\caption{\sgd}
\end{table}

\begin{table}[!htb]
\centering
\setlength{\extrarowheight}{3.5pt}
\begin{tabular}{|c|c|c|c|}
\hline
 & $\eta_0$ & $1-\gamma$ & $\lambda_{L_2}$ \\ \hline
initial & $\srange{e-3}{e1}$ & $\srange{e-3}{e0}$ & $\srange{e-5}{e-2}$ \\ \hline
final & $\srange{e-2}{e0}$ & $\srange{e-1}{e0}$ & $\srange{e-3}{e-1}$ \\ \hline
\end{tabular}
\caption{\momentum}
\end{table}

\begin{table}[!htb]
\centering
\setlength{\extrarowheight}{3.5pt}
\begin{tabular}{|c|c|c|c|c|c|c|}
\hline
 & $\alpha_0 / \sqrt{\epsilon}$ & $1-\gamma$ & $1-\rho$ & $\epsilon$ & $\lambda_{L_2}$\\ \hline
initial & $\srange{e-2}{e4}$ & $\srange{e-3}{e0}$ & \srange{e-3}{e0} & $\srange{e-10}{e10}$ & $\srange{e-5}{e-2}$\\ \hline
final & $\srange{e-2}{e0}$ & $\srange{e-1}{e0}$ & $\srange{e-3}{e-2}$ & $\srange{e2}{e6}$ & $\srange{e-3}{e-1}$\\ \hline
\end{tabular}
\caption{\rmsprop}
\end{table}

\begin{table}[!htb]
\centering
\setlength{\extrarowheight}{3.5pt}
\begin{tabular}{|c|c|c|c|c|c|c|}
\hline
 & $\alpha_0 / \epsilon$ & $1-\beta_1$ & $1-\beta_2$ & $\epsilon$ & $\lambda_{L_2}$\\ \hline
initial & $\srange{e-2}{e4}$ & $\srange{e-3}{e0}$ & $$\srange{e-4}{e-1}$$ & $\srange{e-10}{e10}$ & $\srange{e-5}{e-2}$\\ \hline
final & $\srange{e-2}{e1}$ & $\srange{e-1}{e0}$ & $\srange{e-4}{e-1}$ & $\srange{e6}{e10}$ & $\srange{e-3}{e-1}$\\ \hline
\end{tabular}
\caption{\adam}
\end{table}

\clearpage

\subsection{CNN on CIFAR-100}

\subsubsection{Tuning Constant Learning Rate}\label{appendix:search-space-cnn-cifar100-const-lr}
We fixed all optimizer hyperparameters excluding the learning rate to match those specified in \citet{schneider2019deepobs}.

\begin{table}[!htb]
\centering
\setlength{\extrarowheight}{3.5pt}
\begin{tabular}{|c|c|c|c|}
\hline
& $\eta$ (\sgd) & $\eta$ (\momentum) & $\alpha$ (\adam)\\ \hline
initial & $\srange{e-2}{e0}$ & $\srange{e-4}{e0}$ & $\srange{e-5}{e-2}$ \\ \hline
final & $\srange{e-1}{e0}$ & $\srange{e-3}{e-2}$ & $\srange{e-4}{e-3}$\\ \hline
\end{tabular}
\caption{Learning rate search ranges.}
\end{table}

\subsubsection{Tuning Learning Rate Schedule}\label{appendix:search-space-cnn-cifar100-lr-sched}

We used linear learning rate decay, and tuned the number of decay steps within $\lbrack0.5, 1.0\rbrack$ times the number of training steps and the learning rate decay factor within $\{\num{e-4}, \num{e-3}, \num{e-2}, \num{e-1}\}$.

\begin{table}[!htb]
\centering
\setlength{\extrarowheight}{3.5pt}
\begin{tabular}{|c|c|c|c|}
\hline
& $\eta_0$ (\sgd) & $\eta_0$ (\momentum) & $\alpha_0$ (\adam)\\ \hline
initial & $\srange{e-2}{e0}$ & $\srange{e-4}{e0}$ & $\srange{e-5}{e-2}$ \\ \hline
final & $\srange{e-1}{e0}$ & $\srange{e-3}{e-1}$ & $\srange{e-4}{e-3}$\\ \hline
\end{tabular}
\caption{Learning rate search ranges.}
\end{table}

\subsubsection{Tuning Constant Learning Rate \& \texorpdfstring{$\{\gamma, \beta_1, \beta_2, \epsilon\}$}{Optimizer hyperparameters}}

For \sgd, we reused the results from Appendix~\ref{appendix:search-space-cnn-cifar100-const-lr}, since there were no additional hyperparameters to tune.

\begin{table}[!htb]
\centering
\setlength{\extrarowheight}{3.5pt}
\begin{tabular}{|c|c|c|}
\hline
 & $\eta$ & $1-\gamma$ \\ \hline
final & $\srange{e-4}{e0}$ & $\srange{e-3}{e0}$ \\ \hline
\end{tabular}
\caption{\momentum}
\end{table}

\begin{table}[!htb]
\centering
\setlength{\extrarowheight}{3.5pt}
\begin{tabular}{|c|c|c|c|c|}
\hline
 & $\alpha / \epsilon$ & $1 - \beta_1$ & $1 - \beta_2$ & $\epsilon$ \\ \hline
initial & $\srange{e-2}{e-2}$ & $\srange{e-3}{e0}$ & $\srange{e-4}{e-1}$ & $\srange{e-10}{e-10}$ \\ \hline
final & $\srange{e-1}{e-1}$ & $\srange{e-2}{e0}$ & $\srange{e-4}{e-1}$ & $\srange{e2}{e6}$ \\ \hline
\end{tabular}
\caption{\adam}
\end{table}

\subsubsection{Tuning Learning Rate Schedule \& \texorpdfstring{$\{\gamma, \beta_1, \beta_2, \epsilon\}$}{Optimizer hyperparameters}}

We used linear learning rate decay, and tuned the number of decay steps within $\lbrack0.5, 1.0\rbrack$ times the number of training steps and the learning rate decay factor within $\{\num{e-4}, \num{e-3}, \num{e-2}, \num{e-1}\}$. For \sgd, we reused the results from Appendix~\ref{appendix:search-space-cnn-cifar100-lr-sched}, since there were no additional hyperparameters to tune.

\begin{table}[!htb]
\centering
\setlength{\extrarowheight}{3.5pt}
\begin{tabular}{|c|c|c|}
\hline
 & $\eta_0$ & $1-\gamma$ \\ \hline
initial & $\srange{e-4}{e0}$ & $\srange{e-2}{e0}$ \\ \hline
final & $\srange{e-3}{e-1}$ & $\srange{e-3}{e-1}$ \\ \hline
\end{tabular}
\caption{\momentum}
\end{table}

\begin{table}[!htb]
\centering
\setlength{\extrarowheight}{3.5pt}
\begin{tabular}{|c|c|c|c|c|}
\hline
 & $\alpha_0 / \epsilon$ & $1 - \beta_1$ & $1 - \beta_2$ & $\epsilon$ \\ \hline
initial & $\srange{e-1}{e1}$ & $\srange{e-3}{e0}$ & $\srange{e-4}{e-1}$ & $\srange{e2}{e6}$ \\ \hline
final & $\srange{e-1}{e1}$ & $\srange{e-3}{e0}$ & $\srange{e-5}{e-2}$ & $\srange{e2}{e6}$ \\ \hline
\end{tabular}
\caption{\adam}
\end{table}

\clearpage

\subsection{LSTM on War and Peace}

\subsubsection{Tuning constant Learning Rate}

\begin{table}[!htb]
\centering
\setlength{\extrarowheight}{3.5pt}
\begin{tabular}{|c|c|}
\hline
 & $\eta$ \\ \hline
final & $\srange{e-2}{e1}$ \\ \hline
\end{tabular}
\caption{\sgd}
\end{table}

\begin{table}[!htb]
\centering
\setlength{\extrarowheight}{3.5pt}
\begin{tabular}{|c|c|c|}
\hline
 & $\eta$ & $1-\gamma$ \\ \hline
final & $\srange{e-4}{e0}$ & $0.99$ \\ \hline
\end{tabular}
\caption{\momentum}
\end{table}

\begin{table}[!htb]
\centering
\setlength{\extrarowheight}{3.5pt}
\begin{tabular}{|c|c|c|c|c|}
\hline
 & $\alpha / \epsilon$ & $1-\beta_1$ & $1-\beta_2$ & $\epsilon$ \\ \hline
final & $\srange{e-5}{e-2}$ & $0.9$ & $0.999$ & $\num{e-8}$ \\ \hline
\end{tabular}
\caption{\adam}
\end{table}

\subsubsection{Tuning Learning Rate schedule \& \texorpdfstring{$\{\gamma, \beta_1, \beta_2, \epsilon\}$}{Optimizer hyperparameters}}
\label{sec:search_space_lstm_tolstoy}

We used linear learning rate decay, and tuned the number of decay steps within $\lbrack0.5, 1.0\rbrack$ times the number of training steps and the learning rate decay factor $f$ within the values shown in the tables below.

\begin{table}[!htb]
\centering
\setlength{\extrarowheight}{3.5pt}
\begin{tabular}{|c|c|c|}
\hline
 & $\eta_0$ & $f$ \\ \hline
initial & $\srange{e-3}{e1}$ & $\{\num{e-4}, \num{e-3}, \num{e-2}, \num{e-1}\}$ \\ \hline
final & $\srange{e0}{e1}$ & $\{\num{e-4}, \num{e-3}, \num{e-2}, \num{e-1}\}$ \\ \hline
\end{tabular}
\caption{\sgd}
\end{table}

\begin{table}[!htb]
\centering
\setlength{\extrarowheight}{3.5pt}
\begin{tabular}{|c|c|c|c|}
\hline
 & $\eta_0$ & $1-\gamma$ & $f$ \\ \hline
initial & $\srange{e-4}{e0}$ & $\srange{e-3}{e0}$ & $\{\num{e-4}, \num{e-3}, \num{e-2}, \num{e-1}\}$ \\ \hline
final & $\srange{e-1}{e1}$ & $\srange{e-2}{e0}$ & $\{\num{e-4}, \num{e-3}, \num{e-2}, \num{e-1}\}$ \\ \hline
\end{tabular}
\caption{\momentum}
\end{table}

\begin{table}[!htb]
\centering
\setlength{\extrarowheight}{3.5pt}
\begin{tabular}{|c|c|c|c|c|c|}
\hline
 & $\alpha_0 / \epsilon$ & $1-\beta_1$ & $1-\beta_2$ & $\epsilon$ & $f$\\ \hline
initial & $\srange{e-2}{e4}$ & $\srange{e-3}{e0}$ & $\srange{e-4}{e-1}$ & $\srange{e-10}{e10}$ & $\{\num{e-4}, \num{e-3}, \num{e-2}, \num{e-1}\}$ \\ \hline
final & $\srange{e0}{e2}$ & $\srange{e-2}{e0}$ & 0.999 & $\srange{e0}{e4}$ & $10^{-3}$\\ \hline
\end{tabular}
\caption{\adam}
\end{table}

\clearpage

\section{Additional plots}\label{appendix:extra-plots}

\begin{figure*}[hbt!]
    \centering
    \includegraphics[width=\linewidth]{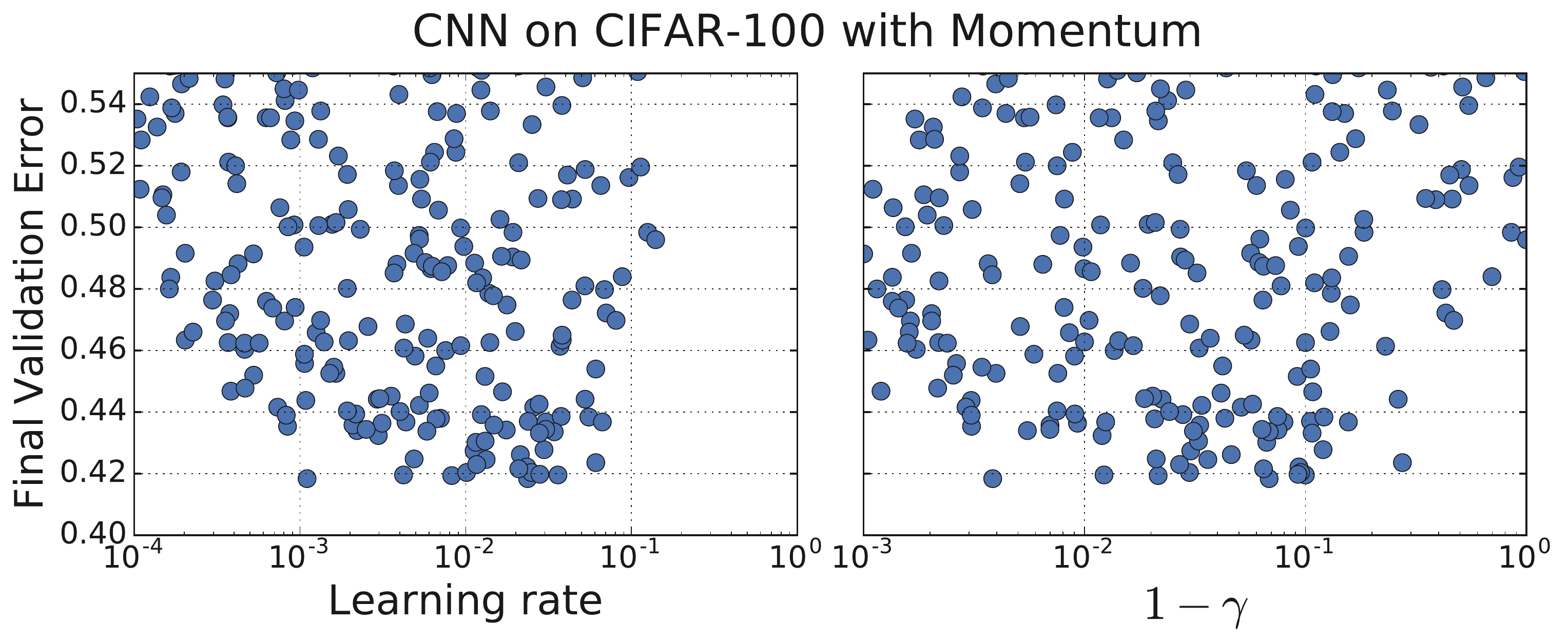}
    \caption{Example plot of final validation error projected onto the axes of the hyperparameter space. We consider this search space to be appropriate because the optimal values are away from the search space boundaries.}
    \label{fig:example_hparam_plot}
\end{figure*}

\begin{figure*}[hbt!]
    \centering
    \includegraphics[width=.9\linewidth]{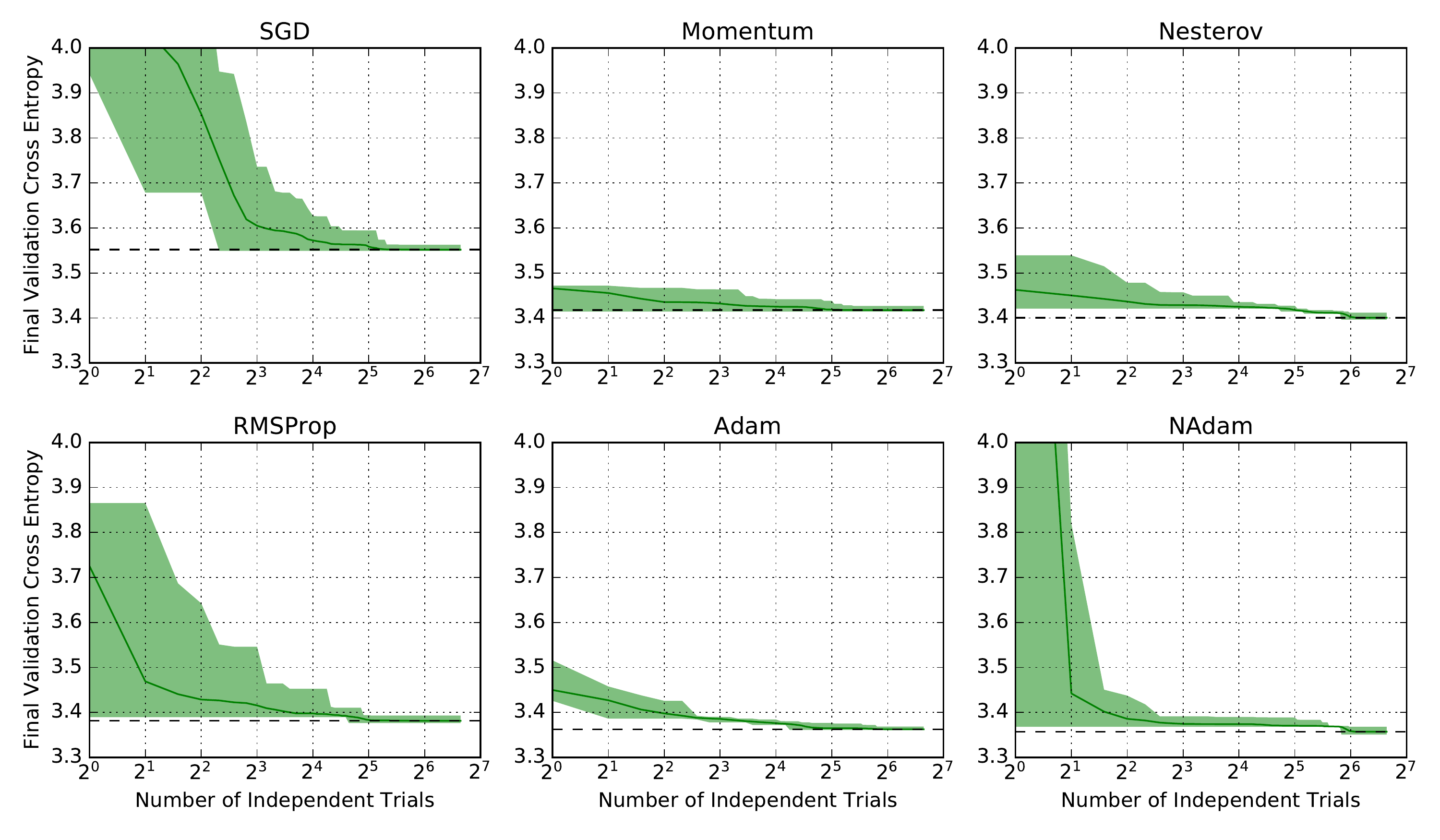}
    \caption{Validation performance of the best trial mostly converges with as few as $2^4$ hyperparameter tuning trials for Transformer on LM1B. Shaded regions indicate $5^\text{th}$ and $95^\text{th}$ percentiles estimated with bootstrap sampling (see Appendix~\ref{appendix:bootstrap}). The search spaces can be found in Appendix~\ref{sec:search_space_transformer_lm1b}.}
    \label{fig:bootstrap-trial-lm1b-transformer}
\end{figure*}

\begin{figure*}[hbt!]
    \centering
    \includegraphics[width=\linewidth]{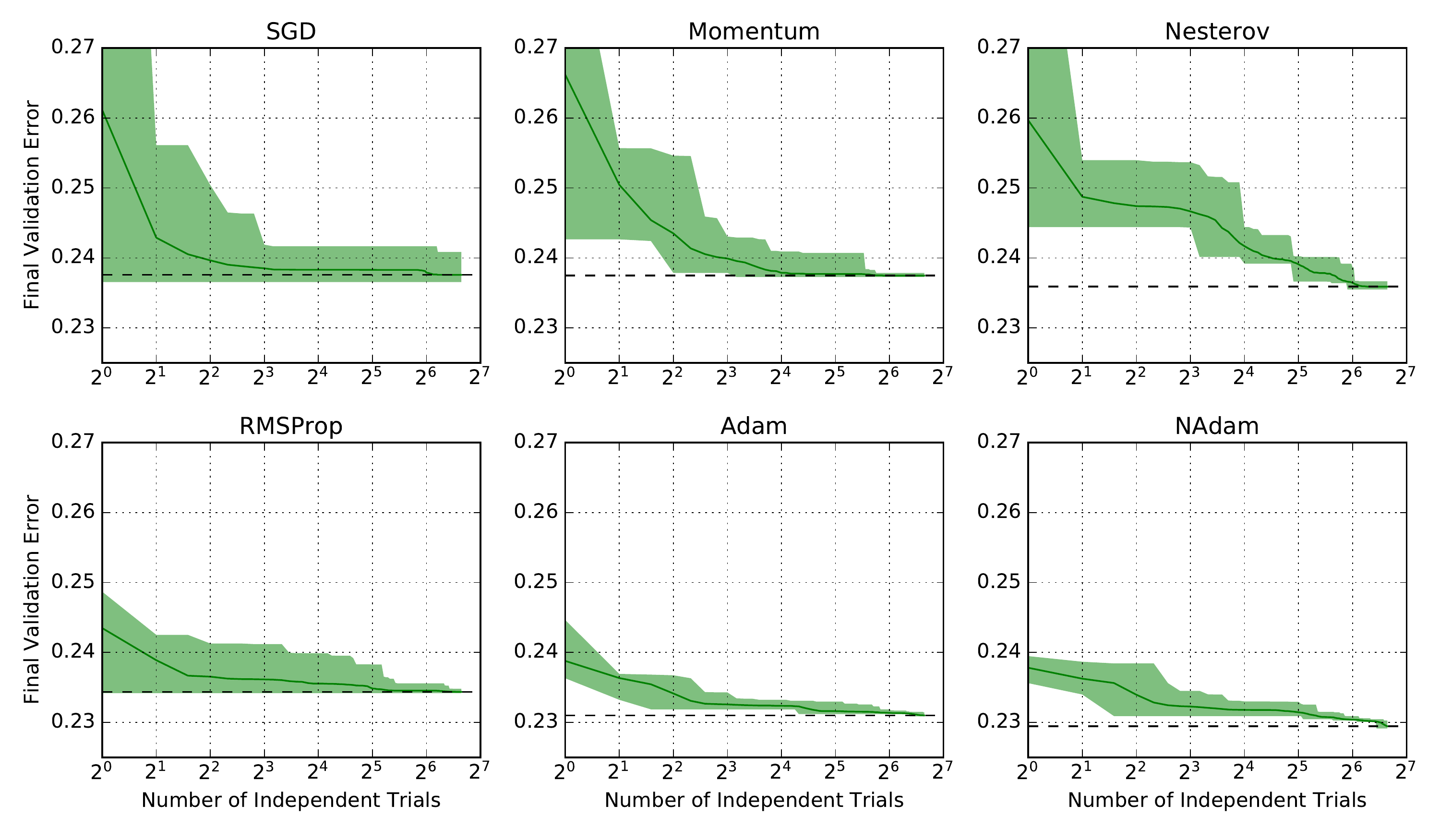}
    \caption{Validation performance of the best trial mostly converges with as few as $2^4$ hyperparameter tuning trials for ResNet-50 in ImageNet. Shaded regions indicate $5^\text{th}$ and $95^\text{th}$ percentile estimated with bootstrap sampling (see Appendix~\ref{appendix:bootstrap}). The search spaces can be found in ~\ref{sec:search_space_resnet_imagenet}.}
    \label{fig:bootstrap-trial-imagenet-resnet}
\end{figure*}

\begin{figure*}[hbt!]
    \centering
    \includegraphics[width=\linewidth]{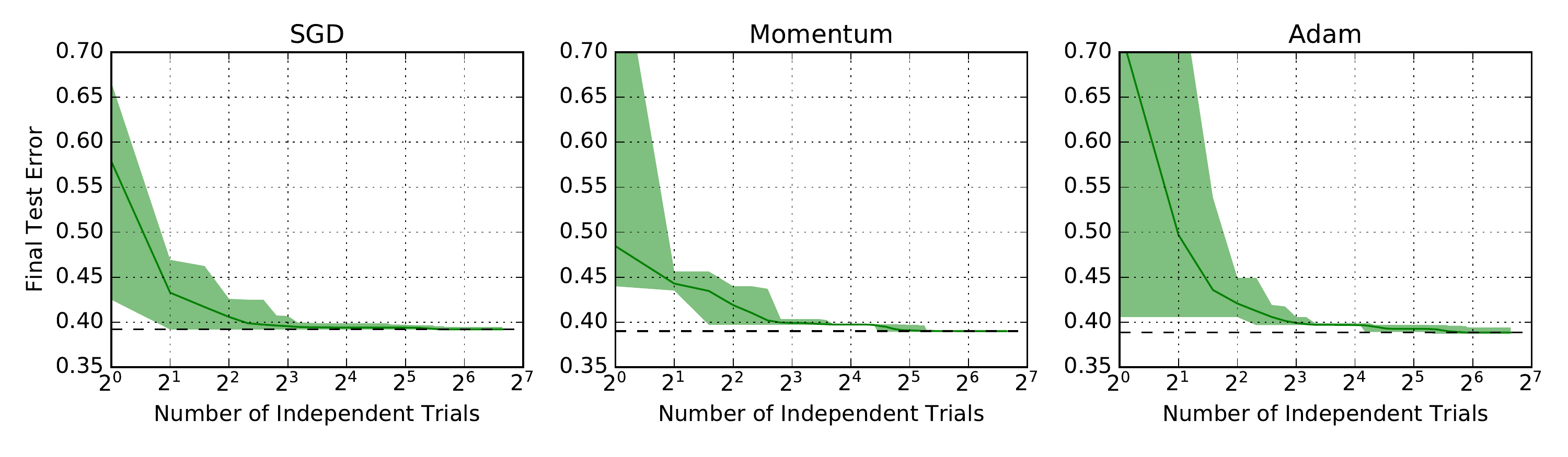}
    \caption{Test performance of the best trial mostly converges with as few as $2^3$ hyperparameter tuning trials for a 2-layer LSTM on \textit{War and Peace}. Shaded regions indicate $5^\text{th}$ and $95^\text{th}$ percentile estimated with bootstrap sampling (see Appendix~\ref{appendix:bootstrap}). The search spaces can be found in ~\ref{sec:search_space_lstm_tolstoy}.}
    \label{fig:bootstrap-trial-tolstoy-lstm}
\end{figure*}

\begin{figure*}[hbt!]
    \centering
    \includegraphics[width=\linewidth]{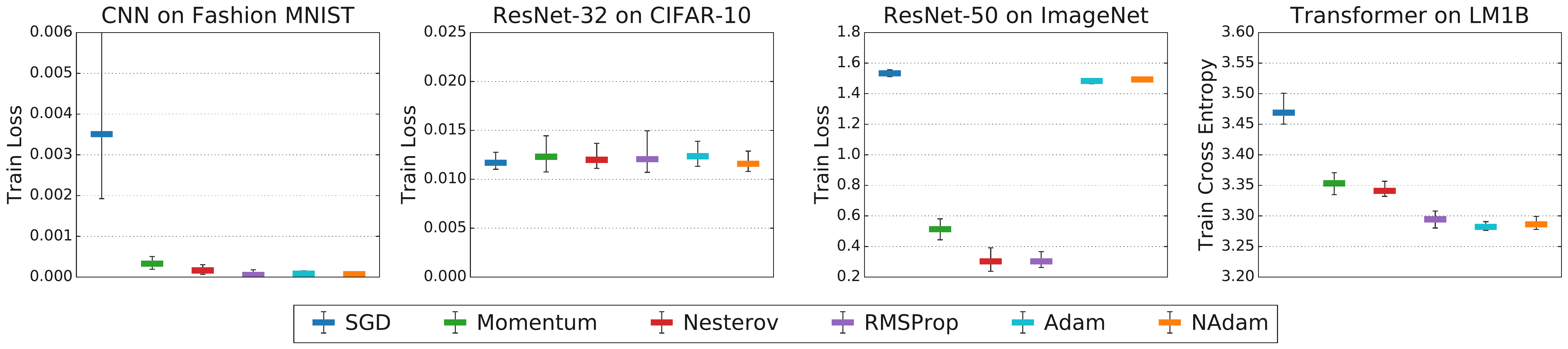}
    \caption{The relative performance of optimizers is consistent with the inclusion relationships when we select for lowest training loss. Note that \sgd{}, \adam{}, and \nadam{} for ResNet-50 on ImageNet used label smoothing in their final search spaces (see Section~\ref{sec:search_space_resnet_imagenet}), which makes their loss values incommensurate with the other optimizers. This is because their final search spaces were optimized to minimize validation error---if we had optimized their search spaces to minimize training error instead, we would not have used label smoothing, and we expect their training loss values would be consistent with the inclusion relationships.}
    \label{fig:our_workloads_train}
\end{figure*}

\begin{figure*}[hbt!]
    \centering
    \includegraphics[width=\linewidth]{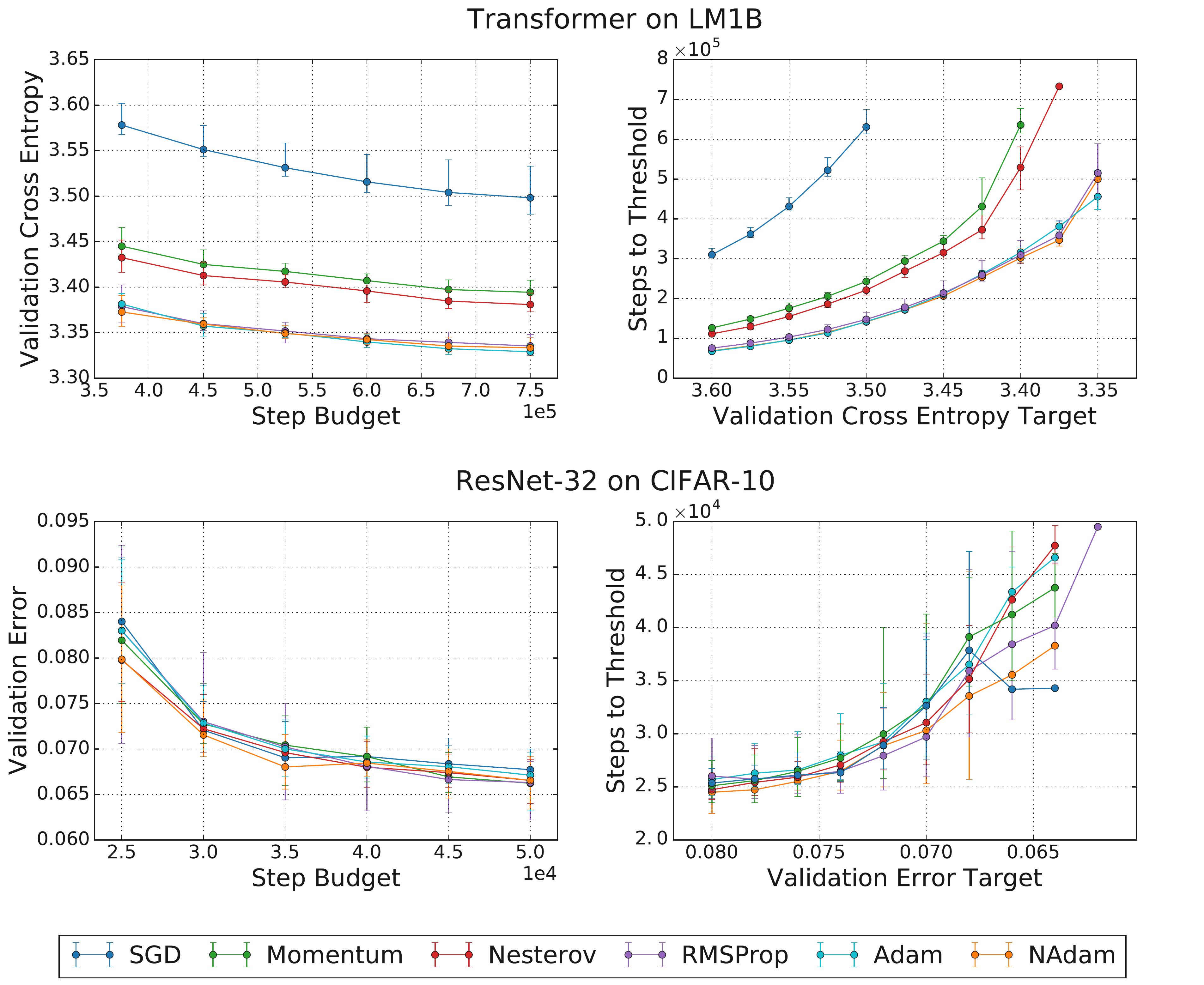}
    \caption{Our results confirming the relevance of optimizer inclusion relationships do not depend on the exact step budgets or error targets we chose.}
    \label{fig:budgets-and-targets}
\end{figure*}

\end{document}